\documentclass[10pt,twocolumn,letterpaper]{article}

\usepackage[pagenumbers]{cvpr}   %

\usepackage[dvipsnames]{xcolor}
\usepackage{multirow}
\usepackage{colortbl}
\usepackage{algorithm2e}
\usepackage{xspace}
\usepackage{graphicx}
\usepackage{overpic}
\usepackage{wrapfig}
\usepackage{makecell}
\usepackage{float}
\usepackage{subfloat}
\usepackage{bm}

\definecolor{lightgray}{gray}{0.9}
\definecolor{mediumgray}{gray}{0.7}

\usepackage{amssymb}%
\usepackage{pifont}%

\newcommand{\revi}[1]{{\color{black}#1}}

\newcommand{\prag}[1]{{\noindent \textbf{#1}}}

\newcommand{\mname}{\emph{PCDreamer}\xspace}

\definecolor{cvprblue}{rgb}{0.21,0.49,0.74}
\usepackage[pagebackref,breaklinks,colorlinks,citecolor=cvprblue]{hyperref}

\title{\mname: Point Cloud Completion Through Multi-view Diffusion Priors}

\author{
    Guangshun Wei$^1$ \quad Yuan Feng$^1$ \quad Long Ma$^1$ \quad Chen Wang$^1$ \quad Yuanfeng Zhou$^{1*}$ \quad Changjian Li$^2$ \vspace{2mm}\\
    $^1$Shandong University \quad $^2$University of Edinburgh \vspace{1mm}\\
    \href{https://gsw-d.github.io/PCDreamer/}{https://gsw-d.github.io/PCDreamer/}
}

\begin{document}

\twocolumn[{
\renewcommand\twocolumn[1][]{#1}
\maketitle 
\begin{center}
\vspace{-6mm}
    \centering
    \begin{overpic}[width=1.0\linewidth]{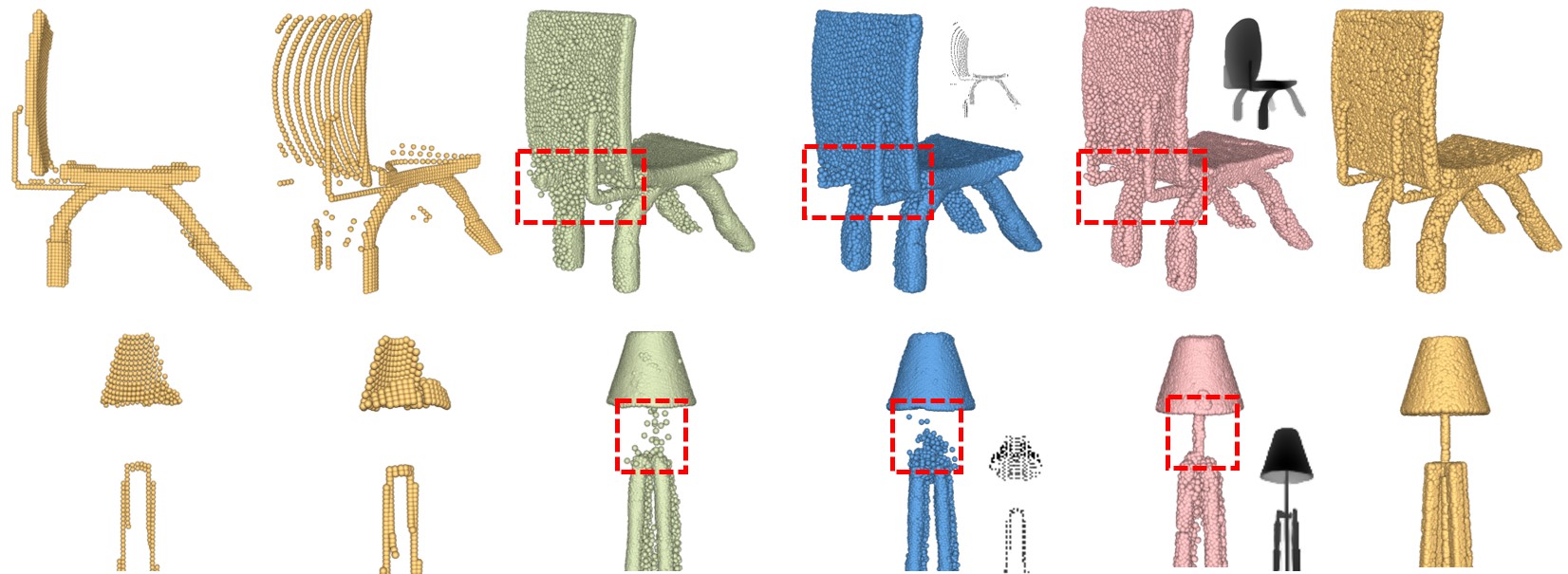} %
        \put(5,-2) {\small (a) Input}
        \put(15,-2) {\small (b) Input of other view}
        \put(35,-2) {\small (c) CRA-PCN~\cite{rong2024cra}}
        \put(53,-2) {\small (d) SVDFormer~\cite{zhu2023svdformer}}
        \put(74,-2) {\small (e) Ours}
        \put(90,-2) {\small (f) GT}
    \end{overpic}
    \captionsetup{type=figure}
    \vspace{-2mm}
    \captionof{figure}{Given a partial point cloud input (a) with (b) as a novel view for visualization purposes, the goal of point cloud completion is to produce a complete point cloud retaining both the global and local geometric features. Existing methods (c, d) fail to neither recover local thin structures (\eg, the lamp holder) nor capture the global symmetric parts (\eg, the back supporter of the chair), while our approach (e) faithfully produces the desired shape compared with the ground truth (f).
    }
    \label{fig:teaser}
\end{center}
}]

\begin{abstract}
\renewcommand{\thefootnote}{}
\footnotetext{$^{*}$Corresponding author.}
This paper presents \mname, a novel method for point cloud completion. 
Traditional methods typically extract features from partial point clouds to predict missing regions, but the large solution space often leads to unsatisfactory results. More recent approaches have started to use images as extra guidance, effectively improving performance, but obtaining paired data of images and partial point clouds is challenging in practice.
To overcome these limitations, we harness the relatively view-consistent multi-view diffusion priors within large models,  to generate novel views of the desired shape. The resulting image set encodes both global and local shape cues, which are especially beneficial for shape completion.
To fully exploit the priors, we have designed a shape fusion module for producing an initial complete shape from multi-modality input (\ie, images and point clouds), and a follow-up shape consolidation module to obtain the final complete shape by discarding unreliable points introduced by the inconsistency from diffusion priors.
Extensive experimental results demonstrate our superior performance, especially in recovering fine details.

\end{abstract}

\section{Introduction}
\label{sec:intro}
Point cloud completion plays a pivotal role in the advancement of 3D vision, serving as a crucial step in numerous applications, including autonomous driving~\cite{reddy2018carfusion, kato2018autoware}, robotics~\cite{cadena2016past, varley2017shape}, and augmented reality~\cite{guo20223d}. Incomplete point clouds, often resulting from scanning occlusions or limited sensor range, pose significant challenges. 
Although many encouraging approaches~\cite{yuan2018pcn, huang2020pf, xie2020grnet, wu2023leveraging, yu2021pointr, xiang2021snowflakenet, yan2022fbnet, tang2022lake, zhou2022seedformer, chen2023anchorformer, zhu2023svdformer, rong2024cra, wen2022pmp, zhang2021view} have been proposed, there is still considerable room for improvement regarding global completeness and local geometric details of resulting point clouds.

This paper addresses this task, focusing on single-view partial point clouds with self-occlusion \cite{yuan2018pcn, huang2020pf, xie2020grnet, yan2022fbnet, yu2021pointr, xiang2021snowflakenet, tang2022lake, wen2022pmp, zhou2022seedformer, chen2023anchorformer, rong2024cra}.
As shown in \cref{fig:teaser}(a) and (b), due to the single-view nature of the scanning process, the self-occlusion often leads to the loss of more than half of the 3D shape. Typical cases, like the lamp without the whole backside and the chair without most of the right side of the seat and legs, are shown in \cref{fig:teaser}. The big missing region implies a vast solution space for completion. 

Most existing methods~\cite{zhu2023svdformer, rong2024cra, zhou2022seedformer, chen2023anchorformer} rely solely on geometric information from partial point clouds to generate a complete shape in a coarse-to-fine manner. 
Although lacking the symmetric supporter of the chair back (\cref{fig:teaser}(c)), the overall chair shape is completed with a certain amount of detail. 
However, the partial lamp without the critical top holder fails those methods, resulting in a random local guess.
To address this, some approaches~\cite{zhang2021view, aiello2022cross, wu2023leveraging, zhu2023svdformer} incorporate additional image information to improve completeness (\cref{fig:teaser}(d), the chair), but obtaining paired image data is challenging in practice.

Motivated by the success of diffusion-based image generation~\cite{long2024wonder3d, liu2024oneplus}, we observe that the relatively view-consistent multi-view images generated by large diffusion models provide rich global and local shape cues, which are particularly advantageous for shape completion. For instance, when conditioned on partial point clouds, the generated images capture symmetric elements, such as the back of the chair (\cref{fig:teaser}(e) top) and the top holder of the lamp (\cref{fig:teaser}(e) bottom), enabling the full shape completion with all relevant features.

To harness the diffusion priors, we propose \mname, a novel algorithm composed of three key components for high-quality point cloud completion. 
Specifically, in the first module, given the single-view partial point cloud, we utilize a large multi-view diffusion model to `dream' a set of realistic images \emph{for free}. These multi-view images depict the desired shape, serving as the fuel for the completion. The second module employs the attention mechanism to effectively fuse both the images and partial point cloud observation, leading to an initial full shape that preserves both the global structure and local features. Additionally, to address the inconsistency introduced by the multi-view images, the final module incorporates a confidence-guided shape consolidator, which produces the final complete shape with high fidelity. 
Extensive evaluation and experiments demonstrate the superior performance of our approach.

\section{Related Work}
\label{ses:rw}

\prag{Partial point cloud-based shape completion.}
Point cloud completion has traditionally relied on partial point clouds, often using intermediate 3D representations processed by 3D convolution, which suffer from resolution limits and high computational costs~\cite{dai2017shape, han2017high, varley2017shape, stutz2018learning}. Point-based networks like PointNet~\cite{qi2017pointnet, qi2017pointnet++} emerged to process 3D coordinates directly, with PCN~\cite{yuan2018pcn} being the first to use an end-to-end point-based approach, generating points through coarse-to-fine folding. Subsequent encoder-decoder methods~\cite{wang2020cascaded, liu2020morphing, xiang2021snowflakenet, wen2022pmp} improved performance.

Recent attention-based approaches, particularly those using vector attention mechanisms~\cite{zhao2021point}, have further advanced the field. SnowflakeNet~\cite{xiang2021snowflakenet} and CRA-PCN~\cite{rong2024cra} introduced novel transformer-based components for improved detail generation. SeedFormer~\cite{zhou2022seedformer} and AnchorFormer~\cite{chen2023anchorformer} both predict shape representations that preserve local information and integrate these representations into the generation process to restore high-fidelity details. Methods like PoinTr~\cite{yu2021pointr} use the partial point cloud as a condition, leveraging Transformer architectures to predict missing points. Despite the improvements, Transformer-based approaches still struggle with severely incomplete point clouds due to limited supplementary information. \revi{SDS-Complete~\cite{kasten2023point} employs semantic priors for point cloud completion, but it suffers from low efficiency and noisy outputs. We compare our method with it qualitatively in the supplementary.}

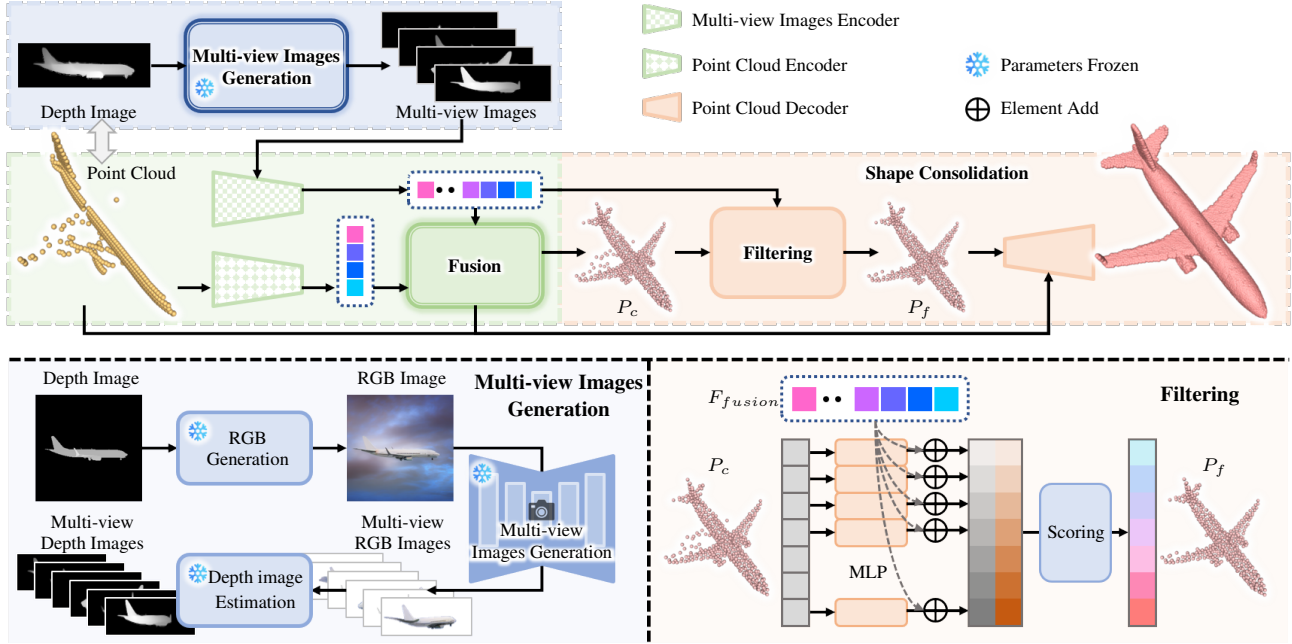
\begin{figure*}[!htb]
    \centering
    \begin{overpic}[width=0.98\linewidth]{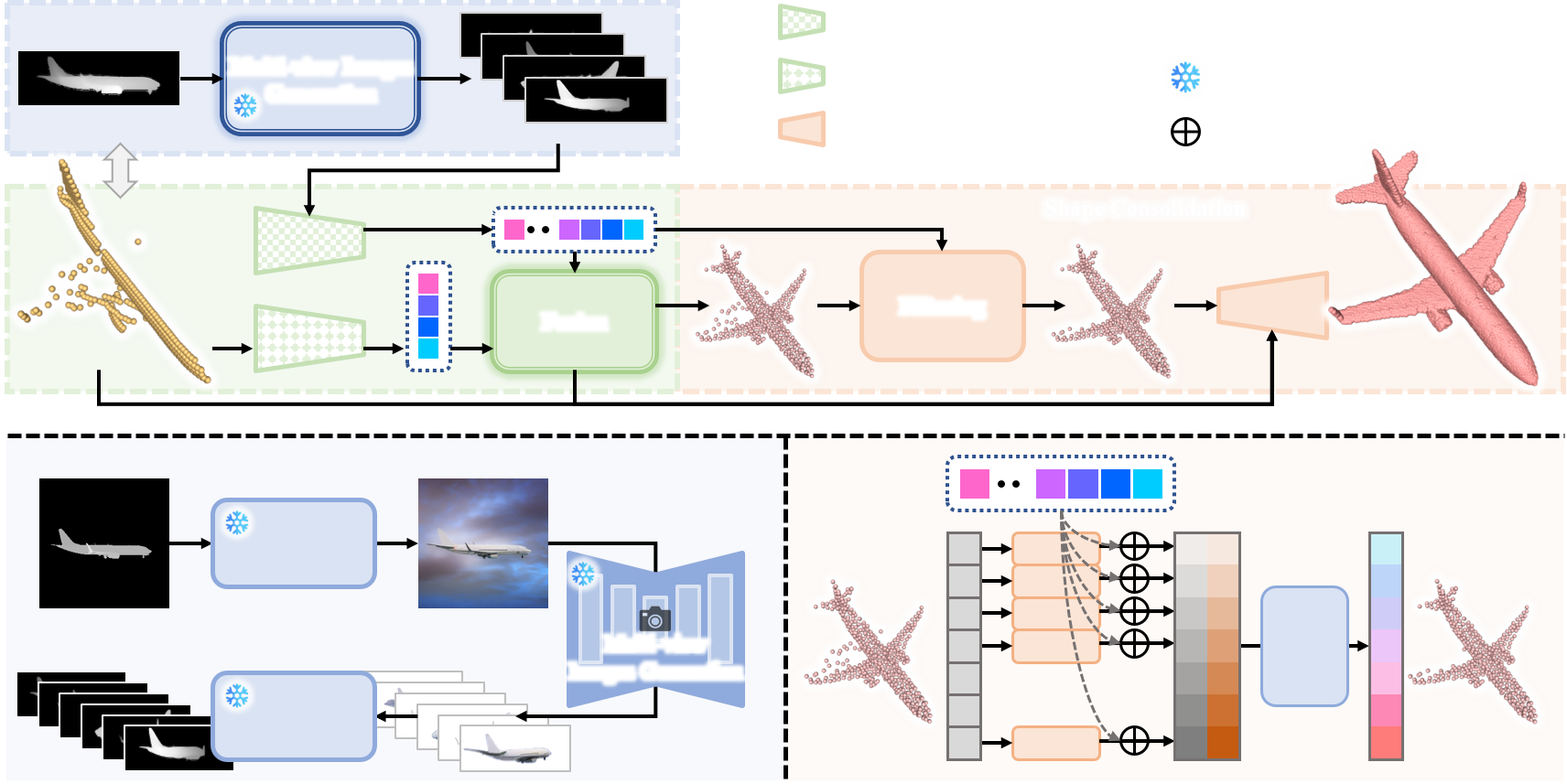}%
        \put(2.9,41){\scriptsize Depth Image}
        \put(30.5,41){\scriptsize Multi-view Images}
        \put(14.7,45.35){\scriptsize \textbf{Multi-view Images}}
        \put(17,43.6){\scriptsize \textbf{Generation}}

        \put(53.5, 48.1){\scriptsize Multi-view Images Encoder}
        \put(53.5, 44.65){\scriptsize Point Cloud Encoder}
        \put(53.5, 41.30){\scriptsize Point Cloud Decoder}
        \put(77.5, 44.65){\scriptsize Parameters Frozen}
        \put(77.5, 41.30){\scriptsize Element Add}
        
        \put(34.55,29){\scriptsize \textbf{Fusion}}
        \put(6.5, 36.2){\scriptsize Point Cloud}
        
        \put(57.5,30.05){\scriptsize \textbf{Filtering}}
        \put(67,36.2){\scriptsize \textbf{Shape Consolidation}}
        \put(47.6,25.8) {\scriptsize $P_c$}
        \put(70.2,25.8) {\scriptsize $P_f$}

        \put(3.1,20.3){\scriptsize Depth Image}
        \put(2.9,7.5){\scriptsize \shortstack{Multi-view\\Depth Images}}
        \put(27.5,20.3){\scriptsize RGB Image}
        \put(27.3,7.5){\scriptsize \shortstack{Multi-view\\RGB Images}}
        \put(15.8,14){\scriptsize \shortstack{RGB\\Generation}}
        \put(15.9,2.9){\scriptsize \shortstack{Depth image\\Estimation}}
        \put(38.6,8.5){\scriptsize Multi-view}
        \put(36.4,6.75){\scriptsize Images Generation}
        \put(36.6, 17.8){\footnotesize \shortstack{\textbf{Multi-view Images}\\\textbf{Generation}}}

        \put(54.7,18.8) {\scriptsize $F_{fusion}$}
        \put(54.7,13.5) {\scriptsize $P_c$}
        \put(93.1,13.5) {\scriptsize $P_f$}
        \put(65.7,5.2) {\scriptsize MLP}
        \put(81.1,8.3) {\scriptsize Scoring}
        \put(89.9, 19){\footnotesize \textbf{Filtering}}
    \end{overpic}
    \vspace{-2mm}
    \caption{\textbf{Overview of \mname}. Given an input partial point cloud, we have designed three core modules to complete it. The multi-view image generation module (\cref{subsec:mv_generation}) \emph{dreams} out multi-view images of the input by leveraging a few large models. The priors within these models serve as the fuel for the completion. The following fusion module (\cref{subsec:fusion}) effectively fuses the original input and the inspiring MV images with the help of the attention mechanism. A final consideration module (\cref{subsec:consolidation}) further reduces the inherent inconsistency introduced in large models, producing a complete, dense, and uniform point cloud with both the global and local shape features.
   }
   \vspace{-3mm}
    \label{fig:pipeline}
\end{figure*}

\prag{Partial point cloud \& Images-based shape completion.}
Given the limited geometric information in partial point clouds, other approaches incorporate additional data like RGB images and depth images to enhance completion performance. MVCN~\cite{hu2019render4completion} uses a conditional GAN for completion in the 2D domain by leveraging rendered depth images from 3D shapes, but this results in spatial information loss. ViPC~\cite{zhang2021view} addresses this by combining partial point clouds with single-view RGB images, though RGB images alone provide insufficient spatial detail. Similarly, Cross-PCC~\cite{wu2023leveraging} employs single-view RGB images and 2D depth images from complete objects, achieving unsupervised point cloud completion without requiring full 3D point clouds. SVDFormer~\cite{zhu2023svdformer} utilizes multi-view depth images from partial point clouds to generate global shapes and produce new points by incorporating learned shape priors and geometric information. However, these depth images, though containing some spatial information, lack comprehensive geometric priors.

\prag{Multi-view images generation.}
Early works~\cite{NEURIPS2023_a0da690a, shi2023mvdream, liu2023syncdreamer} employed specialized attention mechanisms to facilitate cross-view interactions, but often struggled with achieving fine detail consistency and incurred high computational costs. Wonder3D~\cite{long2024wonder3d} tackled these limitations using a diffusion model equipped with cross-domain attention to generate both multi-view images and normal images. More recent approaches~\cite{li2024era3d, huang2024epidiff, hu2024mvd, wang2024crm, qiu2024richdreamer} have enhanced generation quality by incorporating novel attention mechanisms and generating multiple modalities.

Video generation presents even greater challenges due to the need for maintaining temporal consistency. Unlike previous methods~\cite{zhou2022magicvideo, he2023latentvideodiffusionmodels, blattmann2023align, guo2024animatediffanimatepersonalizedtexttoimage, khachatryan2023text2video, zhang2023controlvideo}, Stable Video Diffusion~\cite{blattmann2023stable} introduces a strategy of fine-tuning the entire video diffusion model, rather than focusing solely on temporal layers. This approach provides strong multi-view 3D priors, significantly improving cross-view consistency and generalization, and achieving superior results compared to image-based methods.

\section{Method}
\label{sec:method}

Given a partial, non-uniform point cloud $P_{in} \in \mathbb{R}^{N \times 3}$, our goal is to produce a dense, uniform, and complete point cloud $P_{r} \in \mathbb{R}^{RN \times 3}$ that preserves ac consistent global shape while retaining detailed local features. 
\cref{fig:pipeline} provides an overview of our method consisting of three core components, \ie, multi-view image generation, multi-modality shape fusion, and shape consolidation. In the following, we elaborate on the details.

\subsection{Multi-view Image Generation}
\label{subsec:mv_generation}
Given a partial point cloud observation, we aim to `dream' out the full shape. To this end, we resort to the large multi-view diffusion models. 
There are a few choices for multi-view generation. For instance, RichDreamer~\cite{qiu2024richdreamer} can conditionally generate multi-view depth images. However, the resulting images are of low resolution and poor consistency. To the best of our knowledge, no existing large model is capable of fulfilling this specific purpose. We thus designed a pipeline composed of a few large models in place (\cref{fig:pipeline}, bottom left).

Since most of the multi-view diffusion models take as input an RGB image, we first employ ControlNet~\cite{zhang2023adding} to translate the single-view depth image into an RGB image. We then pass the RGB image to a large diffusion model, for example, Wonder3D~\cite{long2024wonder3d}, which produces six posed multi-view images of the input shape. Note that, our method is not restricted to specific diffusion model, and we have experimented with Wonder3D and Stable Video Diffusion~\cite{blattmann2023stable} (see \cref{tab:Quantitative_results_PCN,tab:Quantitative_results_55,fig:com_shapenet55,fig:pcn}), as a proof of concept. Finally, DepthAnything\cite{yang2024depth} is exploited to generate corresponding depth images for each view, denoted as $I^i_d$ and $I^i_{cam}$ for the depth image and the camera pose, respectively.
The purpose of having depth images is to extract more accurate shape information from the 2.5D images. See our ablation study (\cref{subsec:abl}) on the visual signal selection.

\subsection{Multi-modality Shape Fusion}
\label{subsec:fusion}

Having the multi-view depth images of the imagined shape, next, the key is to effectively fuse them into a full 3D shape that respects the golden partial observation, while accepting the missing regions from multi-view images. We thus designed a multi-modality shape fusion module with two specialized encoders and a fuser.

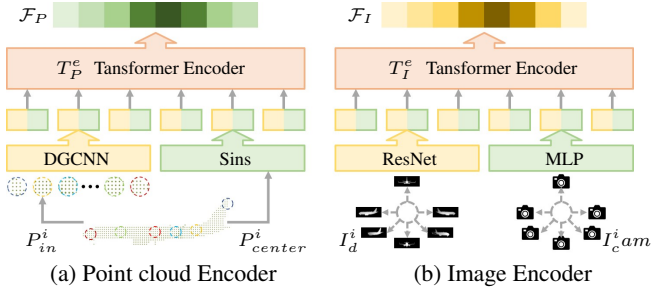
\begin{figure}[!t]
    \centering
    \begin{overpic}[width=\linewidth]{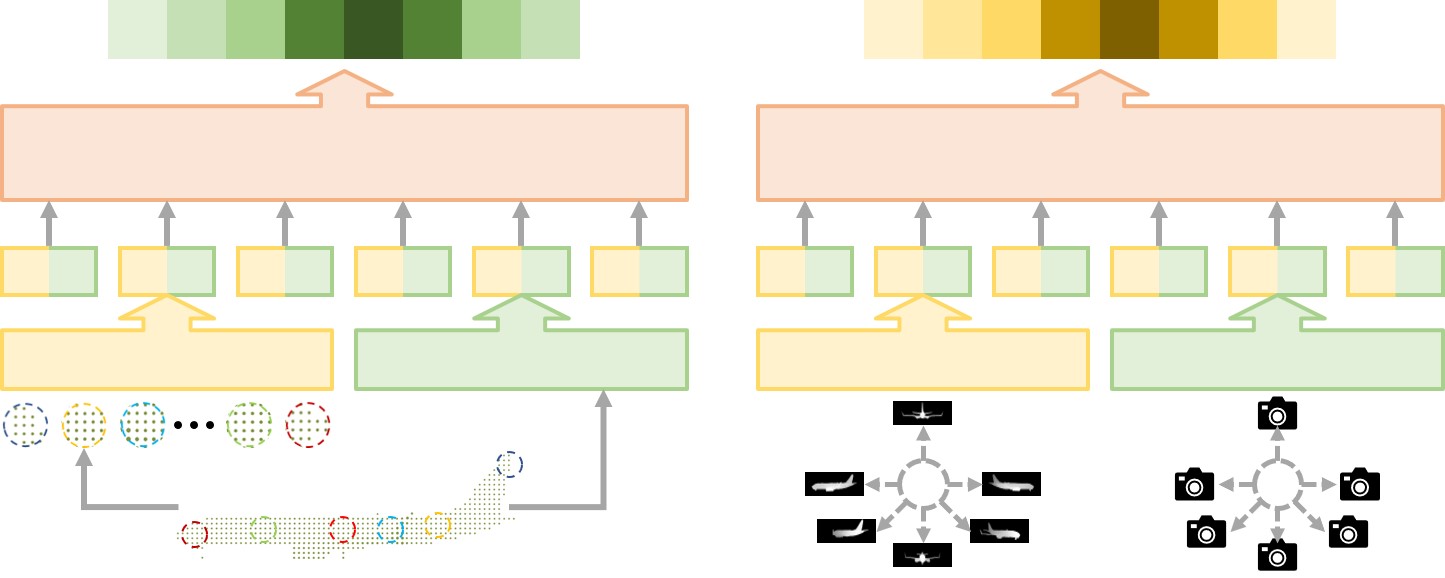}
        \put(7,-6.5) {\small (a) Point cloud Encoder}
        \put(65,-6.5) {\small (b) Image Encoder}
        \put(2,37) {\scriptsize $\mathcal{F}_P$}
        \put(54.5,37) {\scriptsize $\mathcal{F}_I$}
        \put(8,28.5) {\scriptsize $T^e_P$}
        \put(14,28.5) {\scriptsize Tansformer Encoder}
        \put(61,28.5) {\scriptsize  $T^e_I$}
        \put(67,28.5) {\scriptsize Tansformer Encoder}
        \put(6,14) {\scriptsize DGCNN}
        \put(34,14) {\scriptsize Sins}
        \put(3,1) {\scriptsize $P^i_{in}$}
        \put(37,1) {\scriptsize $P^i_{center}$}
        \put(53,1) {\scriptsize  $I^i_d$}
        \put(95,1) {\scriptsize $I^i_cam$}
        \put(60,14) {\scriptsize ResNet}
        \put(86,14) {\scriptsize MLP}
    \end{overpic}  
    \vspace{0.1mm}
    \caption{The partial point cloud and multi-view image encoders. }
    \vspace{-2mm}
    \label{fig:mvEncoder}
\end{figure}

\prag{Partial Point Cloud Encoder.}
Due to the nature of single-view scanning, the partial observation is a golden standard for the desired shape, containing global and local geometric priors. We thus adopted a patch-based Transformer encoder to effectively extract geometric features, as shown in Fig.~\ref{fig:mvEncoder} (a).
Formally, we first select $K$ seed points from the partial point cloud to divide it into patches $P^i_{in}=\{P_1, P_2, \cdots ,P_K\}$, based on a fixed radius $r$. Each patch goes into a DGCNN~\cite{wang2019dynamic} encoder to extract patch-level features, while the center point $P^i_{center}$ (\ie, the seed point) of each patch is transformed into a positional encoding via the sinusoidal function \revi{(denoted as Sins)}. Finally, the Transformer encoder $T^e_P$ takes as input the concatenated features and outputs a unified feature $\mathcal{F}_{P} \in \mathbb{R}^{128}$ of the partial point cloud after average pooling, defined as:
\begin{equation}
    \mathcal{F}_{P} = \text{AvgPool}\left(T^e_{P}(concat(\text{DGCNN}(P^i_{in}), \text{Sins}(P^i_{center}))) \right).
\end{equation}

\prag{Multi-view Image Encoder.}
Given the posed multi-view depth images, a natural way is to fuse them into 3D and extract features from the fused 3D shape directly. We have proved in the ablation study in \cref{subsec:abl} that the resulting shape is irregular and noisy, because of the inherent inconsistency of the multi-view generation. To accurately extract global and local shape cues from the images, particularly for the missing regions, while mitigating the impact of inconsistencies, we have designed a patch-based image encoder, as shown in Fig.~\ref{fig:mvEncoder} (b).
Specifically, each image $I^i_d$ goes through a ResNet backbone to obtain the image feature, while its corresponding camera pose is further encoded via a two-layer MLP to create the positional encoding. Similarly, the Transformer $T^e_I$ takes as input the concatenated feature and produces the final multi-view image fusion feature $\mathcal{F}_I \in \mathbb{R}^{128}$ after average pooling, defined as:
\begin{equation}
    \mathcal{F}_{I} = \text{AvgPool}\left(T^e_{I}\left(concat(\text{ResNet}(I^i_{d}), \text{MLP}(I^i_{cam})) \right )\right).
\end{equation}

\prag{Mutli-modality Fuser.}
In order to effectively exploit the complementary information in both modalities, we resort to the attention mechanism to fuse the aforementioned features. 
Formally, we treat the point cloud feature $\mathcal{F}_{P}$ as the query, while the the image feature $\mathcal{F}_{I}$ as the key and value, defined as:
\begin{equation}
    \bm{Q} = \bm{W}^{Q}{\mathcal{F}}_P, \quad \bm{K} = \bm{W}^K{\mathcal{F}}_I, \quad  \bm{V} = \bm{W}^{V}{\mathcal{F}}_I,
\end{equation}
where the weights are learnable parameters. Then the attention is calculated as follows:
\begin{equation}
    \mathcal{F}_{fusion} = \text{softmax}\left(\frac{\bm{Q}\bm{K}}{\sqrt{128}}\right)\bm{V}.
\end{equation}
A final three-layer MLP converts the fused feature into an initial coarse point cloud $P_c$.

\subsection{Shape Consolidation}
\label{subsec:consolidation}

Due to the inherent inconsistencies in multi-view depth images and limitations in network fitting capabilities, the predicted coarse point cloud tends to be noisy and with localized holes. 
To address these issues, we further propose a confidence-based shape consolidator to produce the final dense completed point cloud.

\prag{Point Filtering.}
As shown in \cref{fig:teaser} bottom right, we assign confidence scores to each point within $P_c$ by examining 1) the consistency between each point and the multi-view feature $\mathcal{F}_I$ and 2) the intra-agreement between points.
Specifically, a point coordinate first goes through an MLP layer to obtain a corresponding feature, which is then concatenated with $\mathcal{F}_I$, as $\mathcal{F}^i_{com}$. We then calculate the average self-attention scores among all the concatenated features:
\begin{equation}
    \begin{split}
            \bm{Q}^i_{com} = \bm{W}^{Q}_{com}{\mathcal{F}}^i_{com}, \quad \bm{K}^i_{com} = \bm{W}^K_{com}{\mathcal{F}}^i_{com}, \\
            S = Sigmoid \left( \text{Avg}\left( dot(\bm{Q}^i_{com}, \bm{K}^i_{com})\right)\right).
    \end{split}
\end{equation}
We empirically divided $P_c$ into high and low confidence sets ($75\%$ vs. $25\%$) based on the calculated confidence score, and take the first set as a filtered intermediate point cloud $P_f$ that is significantly regular and compact. 

\begin{table*}[!htb]
    \centering
    \caption{Quantitative results on the PCN dataset. ($\ell^{1}$ CD ×$10^{3}$ and F-Score@1$\%$). We show the Chamfer distance for each category while reporting the average on the three metrics on the right. `-' indicates the metric is not applicable, and the \colorbox{red!25}{red} and \colorbox{yellow!25}{yellow} colors indicate the best and the second best of the competing methods.}
    \vspace{-2mm}
    \renewcommand{\arraystretch}{1.1}
    \resizebox{0.87\textwidth}{!}{
    \begin{tabular}{r|c c c c c c c c|c c c}
        \toprule
         Methods & Plane & Cabinet & Car & Chair & Lamp & Couch & Table & Boat & CD-Avg$\downarrow$ & DCD-Avg$\downarrow$ & F1$\uparrow$ \\
         \midrule
         PCN~\cite{yuan2018pcn} & 5.82 & 10.91 & 9.00 & 11.09 & 11.91 & 12.06 & 9.56 & 9.12 & 9.93 & - & 0.657 \\
         GRNet~\cite{xie2020grnet} & 8.51 & 12.84 & 10.17 & 12.13 & 11.83 & 12.64 & 11.16 & 9.84 & 11.14 & 0.765 & 0.563 \\
         PoinTr~\cite{yu2021pointr} & 4.31 & 9.23 & 7.60 & 8.35 & 8.27 & 9.38 & 7.97 & 7.01 & 7.76 & 0.562 & 0.810 \\
         PMP-Net++~\cite{wen2022pmp} & 4.80 & 11.70 & 8.96 & 8.93 & 7.04 & 12.25 & 8.57 & 6.88 & 8.64 & 0.714 & 0.683 \\
         SnowFlakeNet~\cite{xiang2021snowflakenet} & 3.95 & 8.82 & 7.52 & 7.48 & 6.34 & 8.88 & 6.61 & 6.10 & 6.96 & 0.567 & 0.828 \\
         SeedFormer~\cite{zhou2022seedformer} & 3.87 & 9.05 & 7.53 & 7.63 & 6.12 & 9.26 & 6.61 & 6.03 & 7.01 & 0.573 & 0.824 \\
         AnchorFormer~\cite{chen2023anchorformer} & 3.62 & 8.79 & 7.20 & 7.12 & 6.18 & 8.71 & 6.50 & 6.01 & 6.77 & 0.539 & 0.841 \\
         HyperCD~\cite{10378095} & 3.87 & 9.05 & 7.40 & 7.50 & 5.95 & 9.25 & 6.50 & 5.89 & 6.93 & \revi{0.568} &\revi{0.822} \\
         SVDFormer~\cite{zhu2023svdformer} & 3.68 & 8.73 & 7.10 & 6.95 & \cellcolor{yellow!25}5.64 & 8.58 & \cellcolor{yellow!25}6.26 & 5.91 & 6.61 & 0.534 & 0.848 \\
         CRA-PCN~\cite{rong2024cra} & 3.62 & 8.77 & 7.00 & 6.92 & \cellcolor{red!25}5.46 & 8.59 & 6.27 & \cellcolor{yellow!25}5.86 & 6.56 & \revi{0.537} & \revi{0.846} \\
         \midrule
         Ours(SVD-MV) & \cellcolor{red!25}3.51 & \cellcolor{red!25}8.62 & \cellcolor{yellow!25}6.92 & \cellcolor{yellow!25}6.91 & 5.66 & \cellcolor{red!25}8.31 & 6.27 & 5.90 & \cellcolor{yellow!25}6.52 & \cellcolor{yellow!25}0.531 & \cellcolor{yellow!25}0.856 \\
         Ours(Wonder3D) & \cellcolor{yellow!25}3.52 & \cellcolor{yellow!25}8.72 & \cellcolor{red!25}6.89 & \cellcolor{red!25}6.71 & \cellcolor{yellow!25}5.64 & \cellcolor{yellow!25}8.32 & \cellcolor{red!25}6.24 & \cellcolor{red!25}5.84 & \cellcolor{red!25}6.49 & \cellcolor{red!25}0.518 & \cellcolor{red!25}0.859 \\
         \bottomrule
    \end{tabular}
    }
    \label{tab:Quantitative_results_PCN}
\end{table*}

\prag{Final Consolidation.}
Firstly, we simply merge $P_f$ and $P_c$ to become a denser point cloud, then apply farthest point sampling (FPS) to obtain a new point set $P_{de} \in \mathbb{R}^{N \times 3}$. We finally upsample $P_{de}$ to produce the dense and uniformly distributed complete point cloud. 

Specifically, we use an MLP to extract per-point shape features ($\mathbb{R}^{128}$) from $P_{de}$ and point-wise concatenate them with the $\mathcal{F}_{fusion}$, resulting in a combine feature of dimension $\mathbb{R}^{256}$. The fused features are then projected to a high-dimensional space to generate a set of offsets $P_\text{offet} \in \mathbb{R}^{RN \times 3}$ (\ie, $R$ times the number of points), defined as: %
\begin{equation}\
\begin{array}{cc}
     &  P_\text{offset}=\text{MLP}\left( \Re \left( concat(\text{MLP}(P_{de}),\mathcal{F}_{fusion})\right)\right),\\
\end{array}
\end{equation}
where $\Re()$ reshapes $ \mathbb{R}^{N \times 256} $ to $ \mathbb{R}^{RN \times 256/R} $. We then map it to $\mathbb{R}^{RN \times 3}$ by a one-layer MLP. Finally, for each point in $P_{de}$, we add the $R$ offsets from $P_\text{offset}$ to produce the final completion result $P_{r} \in \mathbb{R}^{RN \times 3}$.

\paragraph{Network Training}
Our network is trained end-to-end with the Chamfer distance loss. Specifically, we measure the discrepancy between the ground truth point cloud and the coarse ($P_c$) and final ($P_r$) point clouds, respectively. Inspired by \cite{yu2024geoformer}, we exploit the extra hyperbolic Chamfer distance (HyperCD) to combat noise and outliers, defined as:
\begin{equation}
    \mathcal{L}_\text{CD}(X, Y) = \frac{1}{|X|}\sum_{x\in X}\mathop{min}\limits_{y\in Y}||x-y||_{2}^{2} + \frac{1}{|Y|}\sum_{y\in Y}\mathop{min}_{x\in X}||x-y||_{2}^{2}
\end{equation}
and %
\begin{equation}
    \begin{split}
        \mathcal{L}_\text{HCD}(X,Y)=arcosh(1+\mathcal{L}_\text{CD}(X,Y)),
    \end{split}
\end{equation}
where $||x-y||_{2}^{2}$ is the Euclidean distance between point $x$ and $y$. The final loss is:
\begin{equation}
     \mathcal{L}=\mathcal{L}_\text{HCD}(P_c,G)+ \mathcal{L}_\text{HCD}(P_r,G).    
\end{equation}

\begin{figure*}[!htb]
    \centering
    \begin{overpic}[width=\linewidth]{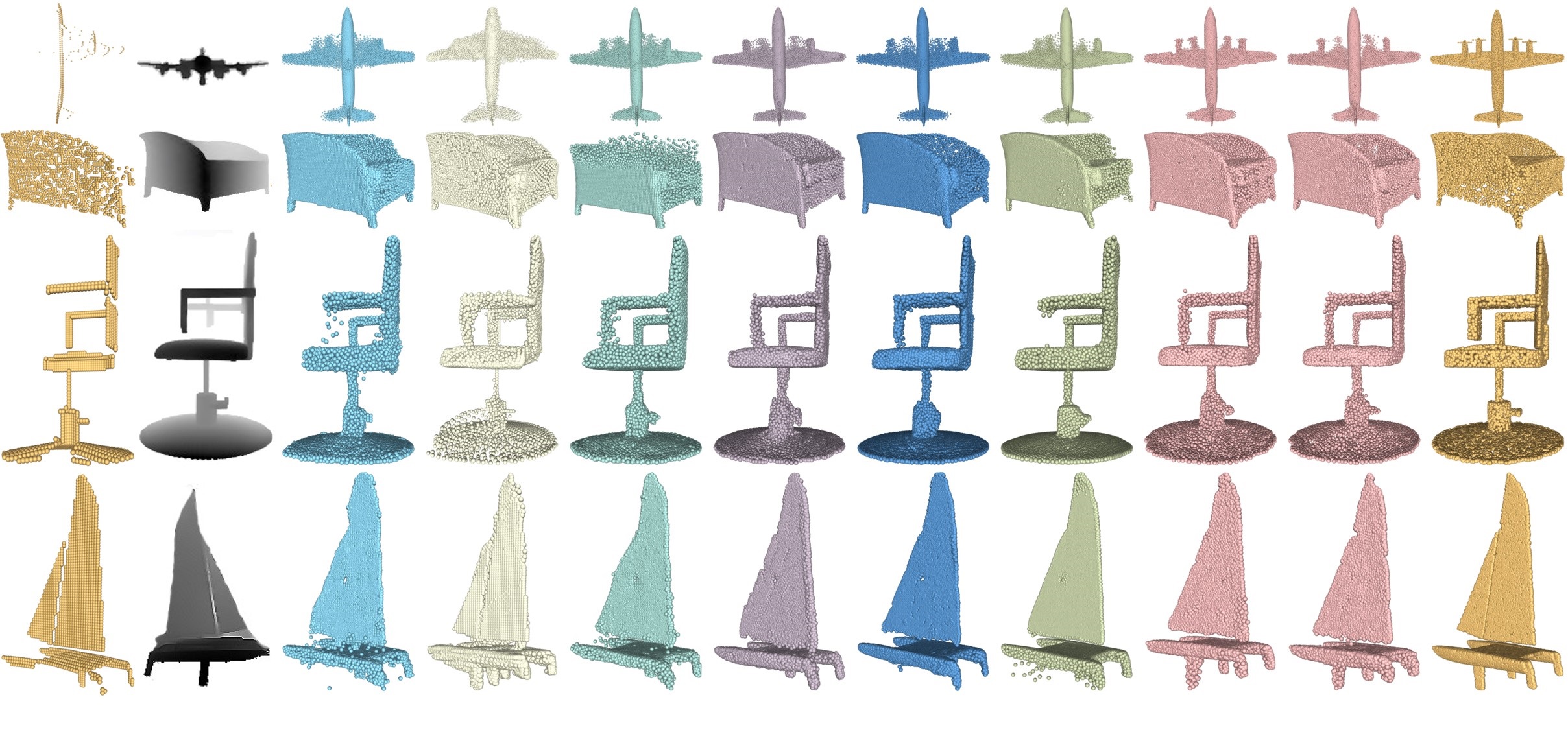}
        \put(3,-0.5) {\scriptsize Input}
        \put(10,-0.5) {\scriptsize Depth image}
        \put(19,-0.5) {\scriptsize SnowFlakeNet}
        \put(30,-0.5) {\scriptsize PoinTr}
        \put(37,-0.5) {\scriptsize SeedFormer}
        \put(45.5,-0.5) {\scriptsize AnchorFormer}
        \put(56,-0.5) {\scriptsize SVDFormer}
        \put(65,-0.5) {\scriptsize CRA-PCN}
        \put(74,-1) {\scriptsize \shortstack{Ours\\(SVD-MV)}}
        \put(83,-1) {\scriptsize \shortstack{Ours\\(Wonder3D)}}
        \put(94,-0.5) {\scriptsize GT}
    \end{overpic}   
    \caption{\textbf{Visual comparisons on the PCN dataset.} That second column is representative depth images selected from the multiple views.}
    \label{fig:pcn}
\end{figure*}

\begin{table*}[!htb]
    \centering
    \caption{Quantitative results on the ShapeNet-55 dataset ($\ell^{2}$ CD ×$10^{3}$ and F-Score@1$\%$). Similarly, Chamfer distance is reported for each category, while the average performance is listed in the right-most columns.}
    \vspace{-2mm}
    \renewcommand{\arraystretch}{1.0}
    \resizebox{\textwidth}{!}{
        \begin{tabular}{r|c c c c c|c c c c c|c c c}
        \toprule
         Methods & Table & Chair & Plane & Car & Sofa & Dishwasher & Mailbox & Microwave & Pillow & Remote & CD-Avg$\downarrow$ & DCD-Avg$\downarrow$ & F1$\uparrow$ \\
         \midrule
         PCN~\cite{yuan2018pcn} & 1.80 & 1.59 & 0.81 & 1.44 & 1.52 & 1.50 & 1.73 & 1.83 & 1.33 & 0.78 & 2.064 & 0.667 & 0.194\\
         GRNet~\cite{xie2020grnet} & 1.72 & 1.54 & 0.82 & 1.56 & 1.61 & 1.77 & 1.39 & 1.91 & 1.39 & 0.79 & 1.801 & 0.696 & 0.243\\
         PoinTr~\cite{yu2021pointr}               & 1.62 & 1.23 & 0.86 & 1.13 & 1.00 & 1.00 & 0.84 & 1.22 & 0.73 & 0.51 & 1.062 & 0.643 & 0.387\\
         AnchorFormer~\cite{chen2023anchorformer} & 1.12 & 0.81 & 0.43 & 1.11 & 0.93 & 1.21 & 0.81 & 1.22 & 0.71 & 0.36 & 1.182 & 0.670 & 0.328\\
         SVDFormer~\cite{zhu2023svdformer}        & 1.03 & 0.80 & 0.44 & 1.00 & \cellcolor{yellow!25}0.88 & 1.00 & 0.76 & 1.27 & 0.71 & \cellcolor{yellow!25}0.34 & 0.938 & 0.621 & \cellcolor{red!25}0.447\\
         CRA-PCN~\cite{rong2024cra}               & 1.58 & 1.08 & 0.58 & 1.24 & 1.19 & 1.31 & 1.18 & 1.31 & 0.86 & 0.45 & 1.300 & \revi{0.0.617} & \revi{0.414}\\
         \midrule
         Ours(SVD-MV) & \cellcolor{yellow!25}0.99 & \cellcolor{yellow!25}0.75 & \cellcolor{yellow!25}0.41 & \cellcolor{yellow!25}0.99 & \cellcolor{yellow!25}0.88 & \cellcolor{red!25}0.94 & \cellcolor{yellow!25}0.61 & \cellcolor{red!25}1.18 & \cellcolor{yellow!25}0.65 & \cellcolor{yellow!25}0.34 & \cellcolor{yellow!25}0.907 & \cellcolor{red!25}0.559 & 0.408\\
         Ours(Wonder3D) & \cellcolor{red!25}0.98 & \cellcolor{red!25}0.71 & \cellcolor{red!25}0.40 & \cellcolor{red!25}0.96 & \cellcolor{red!25}0.85 & \cellcolor{yellow!25}0.98 & \cellcolor{red!25}0.60 & \cellcolor{yellow!25}1.21 & \cellcolor{red!25}0.62 & \cellcolor{red!25}0.32 & \cellcolor{red!25}0.895 & \cellcolor{yellow!25}0.569 & \cellcolor{yellow!25}0.418\\
         \bottomrule
    \end{tabular}
    }
    \label{tab:Quantitative_results_55}
    \vspace{-2mm}
\end{table*}

\section{Experiments and Results}
\label{sec:experiments}

\prag{Implementation.} Our model is trained for 300 epochs on both datasets using four NVIDIA 4090 GPUs with a batch size of 24. We employ the Adam optimizer with an initial learning rate of 1e$^{-4}$. It is decayed by a factor of 0.7 every 40 epochs on the PCN dataset~\cite{yuan2018pcn} and by a factor of 0.98 every 2 epochs on the ShapeNet-55 dataset~\cite{yu2021pointr}. In all of our experiments, we set K to 128 and r to 0.2.
For ControlNet, we set the control strength to 1.0 and the guidance scale to 5.0. For Stable Video Diffusion (SVD-MV), we set the elevation angle to 5 degrees, leveraging its strong generalization capability, while keeping all other parameters at their default values. The resulting video contained 21 frames, from which we selected 6 fixed frames as the multi-view images. Finally, we normalized the generated depth image values to the range of 0 to 1.

\subsection{Dataset}
\label{subsec:Dataset}

\textbf{PCN dataset.} The PCN dataset \cite{yuan2018pcn} is a subset of the ShapeNet dataset \cite{chang2015shapenet}, containing shapes from eight categories. In this study, we constructed our dataset based on PCN. Specifically, given the mesh data of a complete object, we selected a fixed viewpoint along the positive z-axis on its unit sphere to obtain a single-view depth image and generated a partial point cloud with 2,048 points through a back-projection method. \revi{The adoption of a fixed viewpoint is mainly because the large diffusion models (\eg, SVD-MV and Wonder3D) require such an input viewpoint configuration.} Additionally, multi-view images were produced from the single-view depth image using the multi-view image generation module (\cref{subsec:mv_generation}). We uniformly sampled 16,384 points from the mesh surface to obtain the corresponding ground truth point cloud for each object. In total, we collected 30974 samples and split the dataset into training, validation, and test sets, the ratio is the same as \cite{yuan2018pcn}.

\noindent\textbf{ShapeNet-55 dataset.} Unlike the PCN dataset~\cite{yuan2018pcn}, ShapeNet-55~\cite{yu2021pointr} includes shapes from 55 categories
We apply the same construction method as aforementioned to create a dataset.

\noindent\textbf{Evaluation metrics.} Following recent works~\cite{zhou2022seedformer, zhu2023svdformer}, we adopt Chamfer Distance (CD), Density-aware CD (DCD)~\cite{wu2021balanced}, and F1-Score as evaluation metrics, and report the $\ell^{1}$ version of CD for PCN dataset and the $\ell^{2}$ version of CD for Shapenet-55 dataset. 

\subsection{Comparisons}
To demonstrate the effectiveness of our method, we compare it with SoTA methods, including PCN~\cite{yuan2018pcn}, GRNet~\cite{xie2020grnet}, PoinTr~\cite{yu2021pointr}, PMP-Net++~\cite{wen2022pmp}, SnowFlakeNet~\cite{xiang2021snowflakenet}, SeedFormer~\cite{zhou2022seedformer}, AnchorFormer~\cite{chen2023anchorformer}, HyperCD~\cite{10378095}, SVDFormer~\cite{zhu2023svdformer}, and CRA-PCN~\cite{rong2024cra}. We retrain their networks from scratch on our datasets with their default configurations. The quantitative and qualitative results are presented in the following.

\prag{Results on PCN Dataset.} The quantitative results in \cref{tab:Quantitative_results_PCN} demonstrate that our method outperforms all previous approaches across nearly all categories while achieving the best average performance. An exception is the \textit{Lamp}, where CRA-PCN achieves slightly better performance. The inconsistency in the multi-view image generation module, especially when the lamp’s support structure has a folding line or curved shape, leads to lower accuracy in our method.
A visual comparison is presented in \cref{fig:pcn}, where the visual results in several selected categories (\eg, \textit{Plane}, \textit{Chair}, and \textit{Boat}) show that our method can predict the missing parts more accurately due to the additional shape cues provided by the generated multi-view images. 
Particularly, for the \textit{Airplane} category, all other methods fail to complete the missing symmetric parts (\eg, the engines), while our results are more consistent with the ground truth.

\begin{figure}[!htb]
    \centering
    \begin{overpic}[width=\linewidth]{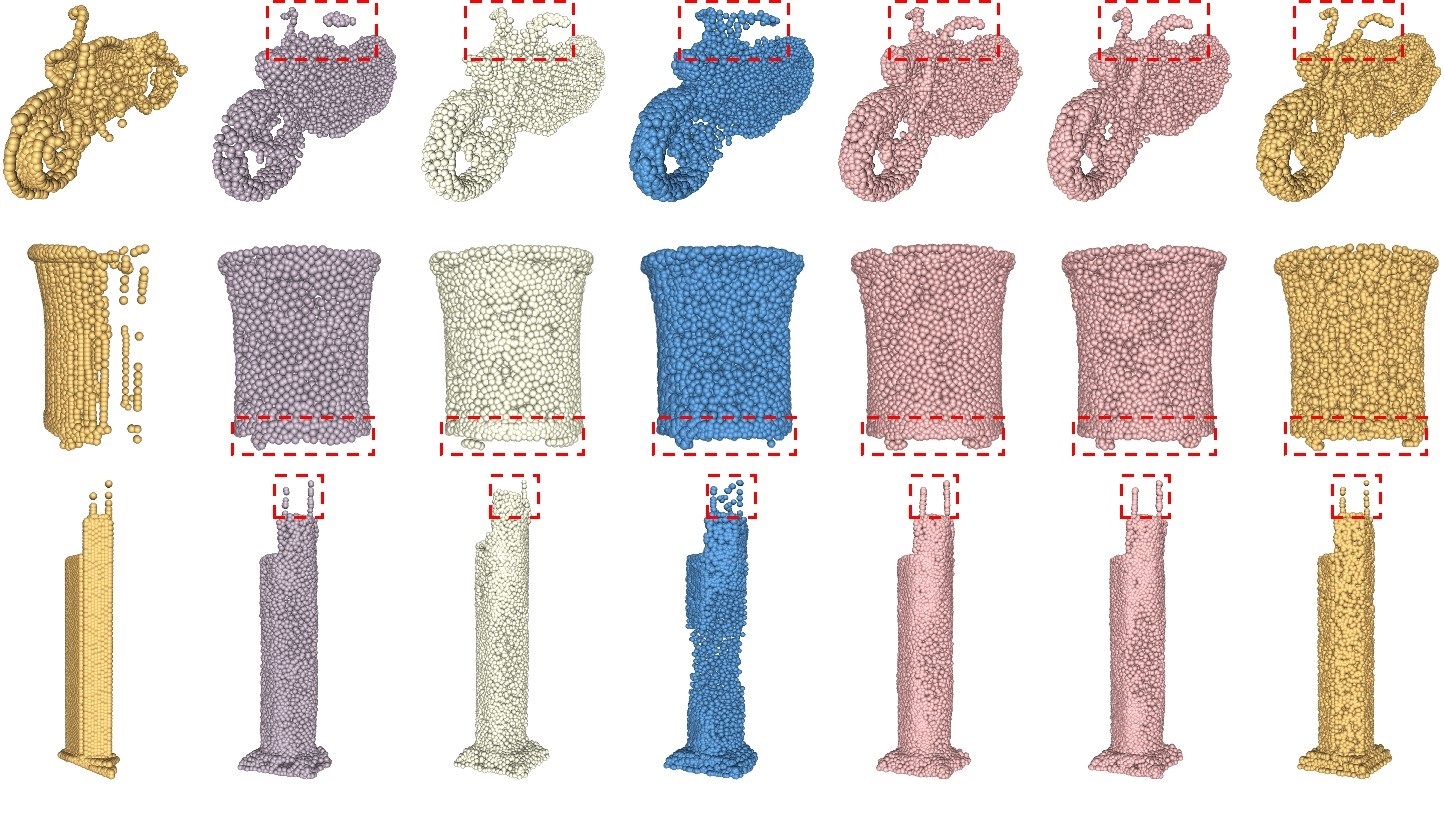}
        \put(3,-1.5) {\scriptsize Input}
        \put(12,-1.5) {\scriptsize AnchorFormer}
        \put(32,-1.5) {\scriptsize PoinTr}
        \put(43,-1.5) {\scriptsize SVDFormer}
        \put(58,-3) {\scriptsize \shortstack{Ours\\(SVD-MV)}}
        \put(72,-3) {\scriptsize \shortstack{Ours\\(Wonder3D)}}
        \put(90,-1.5) {\scriptsize GT}
    \end{overpic}   
    \caption{\textbf{Visual comparisons on the ShapeNet-55 dataset.} The red-boxed regions highlight the effectiveness of different methods in completing local structures.}
    \vspace{-4mm}
    \label{fig:com_shapenet55}
\end{figure}
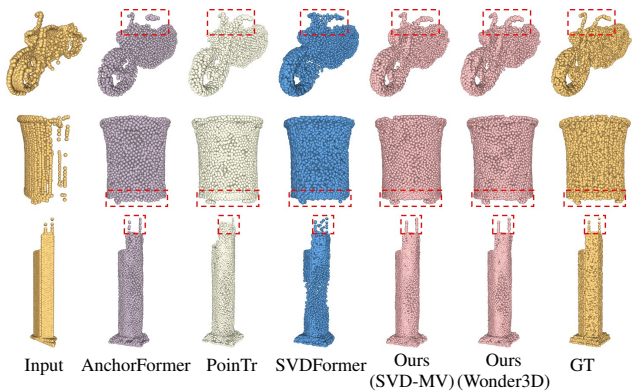

\noindent\textbf{Results on the ShapeNet-55 Dataset.} Following existing work~\cite{zhu2023svdformer}, we also compare our method against some competitors~\cite{yuan2018pcn, xie2020grnet, yu2021pointr, chen2023anchorformer, zhu2023svdformer, rong2024cra} on the ShapeNet-55 dataset.
We did not compare with SeedFormer-based methods~\cite{zhou2022seedformer, 10378095} because we have observed occasional numerical failure during their training caused by the occurrence of NaN values.
As shown in \cref{tab:Quantitative_results_55}, we selected 10 representative classes from a total of 55 categories to showcase the Chamfer Distance (CD). 
Our method consistently outperforms other state-of-the-art approaches across all categories, with a particularly notable improvement of 0.16 in the CD metric for the \textit{Mailbox} category. 
Although the first five classes contain significantly more samples than the latter five, we have obtained comparable results across the ten categories, which demonstrates the generalization ability and robustness of our method on diverse types of data.
The visual comparison in \cref{fig:com_shapenet55} also clearly illustrates the superior performance on more complex shapes with finer details.

\revi{
\begin{table}[!tb]
    \centering
    \caption{Quantitative results on the ShapeNet-34 dataset ($\ell^{2}$ CD ×$10^{3}$ and F-Score@1$\%$).}
    \vspace{-3mm}
    \scalebox{0.8}{
        \begin{tabular}{r|c c c|c c c}
        \hline \multirow[c]{2}{*}{Methods} & \multicolumn{3}{|c|}{34 seen categories} & \multicolumn{3}{|c}{21 unseen categories} \\
          & CD$\downarrow$ & DCD$\downarrow$ & F1$\uparrow$ & CD$\downarrow$ & DCD$\downarrow$ & F1$\uparrow$ \\
         \midrule
         SVDFormer      &0.839 &0.625&\cellcolor{yellow!25}0.402&1.129&0.617&\cellcolor{yellow!25}0.348\\
         CRA-PCN           & 0.818&0.591&0.387&1.277&0.601& 0.317\\
         \midrule
         Ours(SVD-MV) &\cellcolor{yellow!25}0.765 & \cellcolor{red!25}0.562&0.396& \cellcolor{red!25}{1.040} & \cellcolor{red!25}{0.574}&0.339\\
         Ours(Wonder3D) & \cellcolor{red!25}{0.758}& \cellcolor{yellow!25}{0.573}&\cellcolor{red!25}{0.403}&\cellcolor{yellow!25}1.094& \cellcolor{yellow!25}0.582& \cellcolor{red!25}{0.352}\\
         \bottomrule
    \end{tabular}
    }
    \vspace{-2mm}
    \label{tab:Quantitative_results_34}
\end{table}
}

\revi{\noindent\textbf{Results on the ShapeNet-34 Dataset.} To further evaluate the performance of our method on unseen object categories, we also conduct experiments on the ShapeNet-34 dataset~\cite{yu2021pointr}. \cref{tab:Quantitative_results_34} displays the statistical results, where \mname maintains high completion quality on unseen categories (see the CD and DCD metrics) compared against state-of-the-art methods~\cite{zhu2023svdformer, rong2024cra}, demonstrating moderately strong generalization ability.
Note that we process the data as stated in \cref{subsec:Dataset}.
}

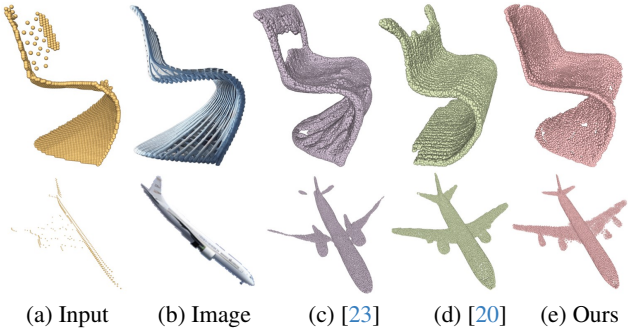
\begin{figure}[!htb]
    \centering
    \begin{overpic}[width=\linewidth]{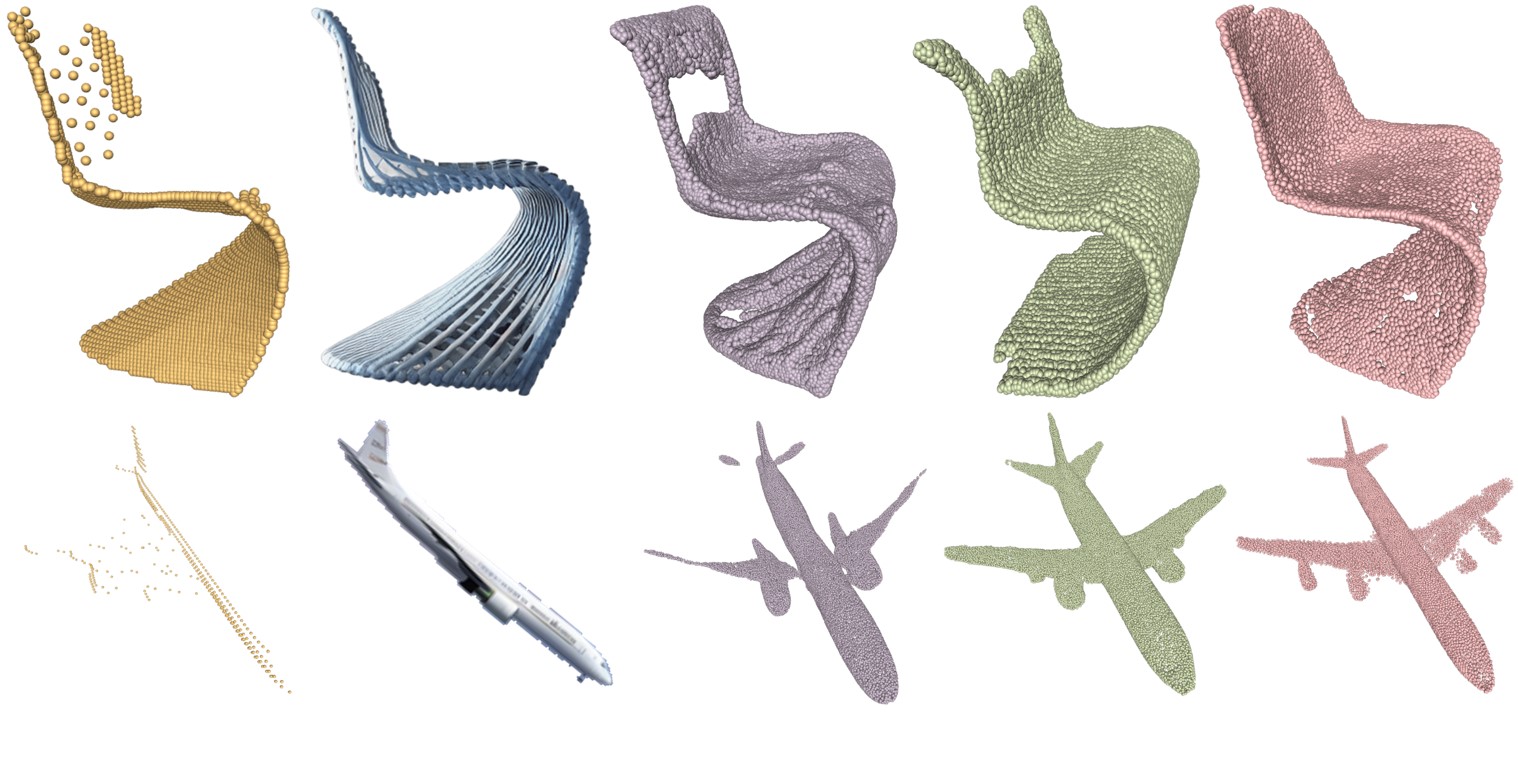}
        \put(4,-0.0) {\small (a) Input}
        \put(23,-0.0) {\small (b) Image}
        \put(45,-0.0) {\small (c) \cite{liu2024oneplus}}
        \put(66,-0.0) {\small (d) \cite{li2024craftsman}}
        \put(86,-0.0) {\small (e) Ours}
    \end{overpic}   
    \vspace{-6mm}
    \caption{\textbf{Visual comparisons with 3D model generation baesd on single-view RGB image.} (b) shows an RGB image generated by ControlNet from a single-view depth image. (c) One-2-3-45++~\cite{liu2024oneplus}, (d) Crafsman~\cite{li2024craftsman}.   }
    \label{fig:comparison_recon}
    \vspace{-4mm}
\end{figure}

\paragraph{Comparison with 3D model generation.} We further compare our approach with recent methods~\cite{liu2024oneplus, li2024craftsman} that are pre-trained on large-scale datasets and leverage powerful diffusion models to generate complete 3D models from a single-view RGB image, thereby enabling effective shape completion as well. 
For a fair comparison, we employ ControlNet to generate an RGB image from a single-view depth image, which is then used as input to their pretrained models to predict the complete 3D shape. 
Finally, the point cloud is extracted using Poisson disk sampling. As shown in Fig.~\ref{fig:comparison_recon}, although One-2-3-45++~\cite{liu2024oneplus} and Crafsman~\cite{li2024craftsman} can generate overall regular shapes, they fail to capture local geometric features, \eg, the back of the chair and the symmetric engines of the airplane, due to the limited cues from the single-view image. 
Instead, our methods can faithfully recover a full shape with fine details based on the cues from the multi-view images. More comparison results can be found in the supplementary.

\subsection{Ablation study}
\label{subsec:abl}

We have extensively ablated our core design choices, and report both the qualitative and quantitative results on the PCN dataset in the following.

\begin{table}[!tb]
    \centering
    \caption{Effect of applying both partial point cloud and depth images ($\ell^{1}$ CD ×$10^{3}$ and F-Score@1$\%$).}
    \vspace{-3mm}
    \renewcommand{\arraystretch}{1.0}
    \resizebox{0.8\linewidth}{!}{
        \begin{tabular}{c c |c c c}
        \toprule
         \makecell[c]{Depth\\images} & \makecell[c]{Partial\\input} & CD-Avg$\downarrow$ & DCD-Avg$\downarrow$ & F1$\uparrow$ \\
         \midrule
         \checkmark & & 17.69 & 0.823 & 0.401 \\
          & \checkmark & 6.91 & 0.534 & 0.840 \\
         \checkmark & \checkmark & \textbf{6.49} & \textbf{0.518} & \textbf{0.858} \\
         \bottomrule
        \end{tabular}
    }
    \label{tab:ablation_depth_coord}
    \vspace{-2mm}
\end{table}

\begin{table}[!tb]
    \centering
    \caption{Ablation on the type of multi-view images ($\ell^{1}$ CD ×$10^{3}$ and F-Score@1$\%$).}
    \vspace{-3mm}
    \renewcommand{\arraystretch}{1.0}
    \resizebox{0.8\linewidth}{!}{
        \begin{tabular}{c c| c c c}
        \toprule
         \makecell[c]{RGB\\images} & \makecell[c]{depth\\images} & CD-Avg$\downarrow$ & DCD-Avg$\downarrow$ & F1$\uparrow$ \\
         \midrule
         \checkmark & & 6.63 & 0.530 & 0.851 \\
           &\checkmark & \textbf{6.49} & \textbf{0.518} & \textbf{0.858} \\
         \checkmark & \checkmark & 6.59 & 0.535 & 0.851 \\
         \bottomrule
        \end{tabular}
    }
    \label{tab:ablation_depth_rgb}
    \vspace{-4mm}
\end{table}

\prag{Is depth image a good choice?}
From the multi-view image generation module, we obtained both the depth and RGB images. The former contains 2.5D information of the underlying geometry, while the latter presents more vivid appearance cues. 
In this test, we use either the RGB image, the depth image, or their combination as our multi-view signal in the later modules in our pipeline, and present the statistical results in \cref{tab:ablation_depth_rgb}.   
As can be seen, the depth-only option outperforms others across multiple metrics. The main reason is that the RGB images generated by ControlNet occasionally contain unrealistic artifacts or inconsistencies, which can adversely affect the completion results.

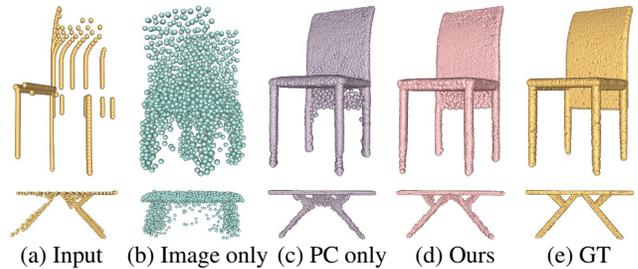
\begin{figure}[h]
    \centering
    \begin{overpic}[width=\linewidth]{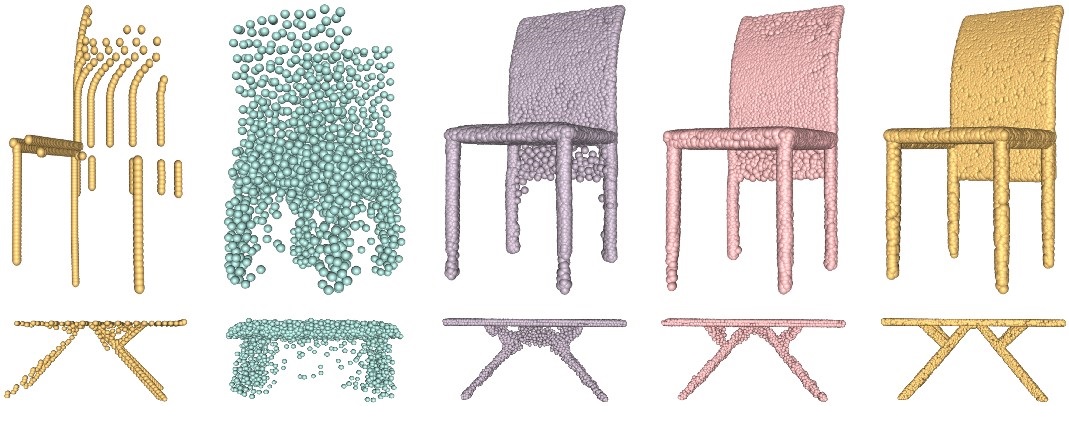} %
        \put(2,-2.2) {\small (a) Input}
        \put(19,-2.2) {\small (b) Image only}
        \put(43,-2.2) {\small (c) PC only}
        \put(65,-2.2) {\small (d) Ours}
        \put(86,-2.2) {\small (e) GT}
    \end{overpic}
    \caption{\textbf{Visual results of multi-modality fusion.} (b) uses only multi-view images, and (c) uses only the partial point cloud. }
    \label{fig:comparison_rgb}
\end{figure}

\prag{Does multi-modality fusion play an important role?}
To evaluate the effectiveness of our multi-modality fusion strategy, we split the network architecture into two branches. The first branch generates a complete point cloud directly from multi-view images, \ie, after getting $\mathcal{F}_I$ from $T_I^e$, we use an MLP to map it to a coarse point cloud, which is then refined by the decoder module to produce the final result. The second branch operates solely on the partial point cloud by removing the multi-view image module from our pipeline. 
Quantitative results of ~\cref{tab:ablation_depth_coord} demonstrate that the multi-modality feature fusion strategy achieves the best performance. As shown in \cref{fig:comparison_rgb}, incorporating additional information from multi-view images allows the model to better capture both the global shape and local structures, such as the backrest of a chair and the supporting regions of a table.

\begin{table}[!tb]
    \centering
    \vspace{-2mm}
    \caption{Effect of confidence filtering ($\ell^{1}$ CD ×$10^{3}$ and F-Score@1$\%$).}
    \vspace{-2mm}
    \renewcommand{\arraystretch}{1.0}
    \resizebox{0.85\linewidth}{!}{
        \begin{tabular}{c|c c c}
        \toprule
         Methods & CD-Avg$\downarrow$ & DCD-Avg$\downarrow$ & F1$\uparrow$ \\
         \midrule
         \makecell[c]{w/o confidence filtering} & 6.76 & 0.525 & 0.845 \\
         Ours & \textbf{6.49} & \textbf{0.518} & \textbf{0.858} \\
         \bottomrule
        \end{tabular}
    }
    \label{tab:ablation_confidence_filtering}
    \vspace{-1mm}
\end{table}

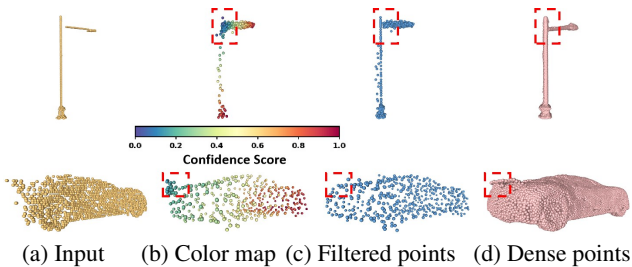
\begin{figure}[!tb]
    \centering
    \begin{overpic}[width=\linewidth]{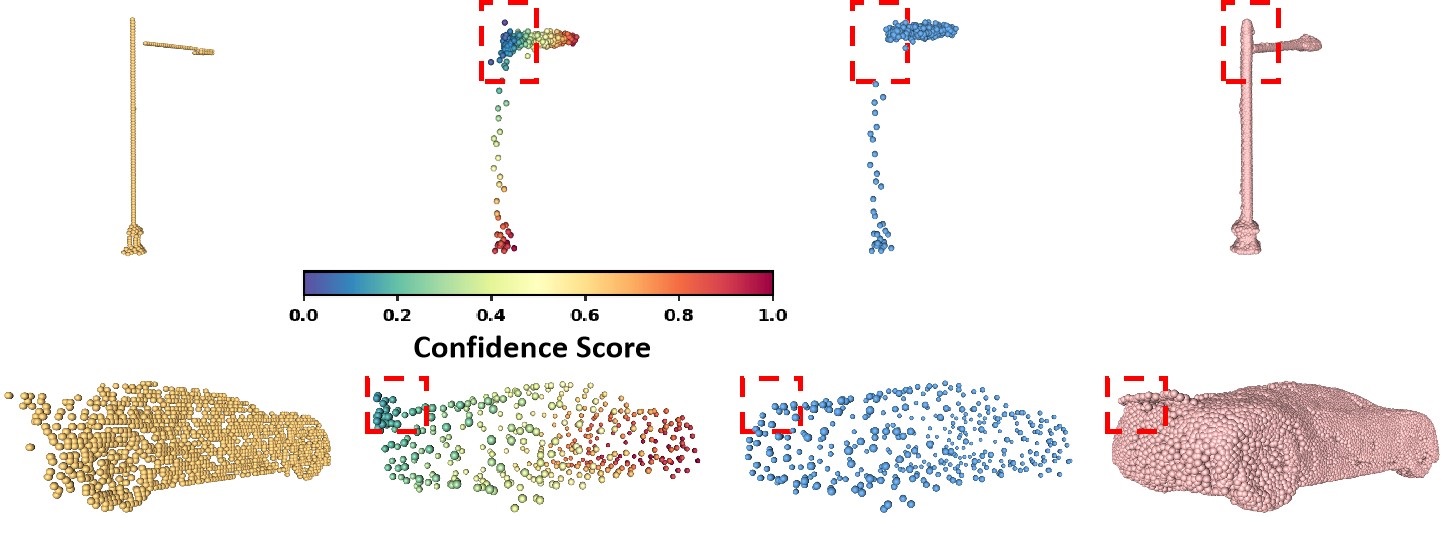}%
        \put(7,-2) {\scriptsize(a) Input}
        \put(27,-2) {\scriptsize (b) Color map}
        \put(49,-2) {\scriptsize (c) Filtered points}
        \put(78,-2) {\scriptsize (d) Dense points}
    \end{overpic}
    \caption{\textbf{Visualization of confidence filtering.}}
    \vspace{-3mm}
    \label{fig:confidenc_filtering}
\end{figure}

\prag{Is confidence filtering helpful?}
To evaluate the impact of our proposed confidence-based filtering operation, we conducted an ablation study by removing it from our method. As shown in   \cref{fig:confidenc_filtering}, the red box in (b) highlights noisy points that are assigned low confidence scores by our method. By applying the filtering strategy, as shown in (c), these noisy points are effectively eliminated. \revi{The remaining points with high confidence are fused with the input and are further consolidated to obtain an} improved completion result (\cref{fig:confidenc_filtering}(d)). Quantitative results in \cref{tab:ablation_confidence_filtering} further demonstrate the effectiveness of the confidence filtering operation.

\subsection{Discussions}

\revi{In the following, we primarily discuss the impact of the number of views, the hyperparameter configuration of filtering, as well as the limitation, while leaving more discussions on handling real-world \emph{LiDAR scans} and \emph{general partial point clouds} in the supplementary.}

\prag{Number of views.}
The guidance provided by multi-view depth images is crucial for point cloud completion, and the number of views $V$ is a key hyperparameter that significantly affects the results. To investigate, we have conducted a series of experiments. First, the SVD-MV model generates 21 frames for each object. We use this as a candidate set and randomly select $V=2, 4, 6, 8$ views from it to train the model separately. 
As shown in \cref{tab:ablation_view_nums}, fewer views result in limited coverage of the object's overall shape. As $V$ increases, the entire shape becomes fully covered, and further increment of $V$ only leads to larger network parameters without significantly improving performance.

\prag{Hyperparameters of filtering.} The filtering strategy can effectively reduce noise in the coarse point cloud, thereby improving the accuracy of completion. However, the number of points filtered needs to be carefully balanced. As shown in \cref{tab:ablation_filtered_points}, filtering too many points can lead to a loss of shape features, affecting the final accuracy, while filtering too few points may retain noise, also impacting precision. 
To address this, we conducted experiments with different filtering levels and selected the optimal setting.

\begin{table}[!tb]
    \centering
    \caption{Number of views ($\ell^{1}$ CD ×$10^{3}$ and F-Score@1$\%$).}
    \vspace{-2mm}
    \renewcommand{\arraystretch}{1.0}
    \resizebox{0.7\linewidth}{!}{
    \begin{tabular}{c| c c c}
        \toprule
             View number & CD-Avg$\downarrow$ & DCD-Avg$\downarrow$ & F1$\uparrow$ \\
             \midrule
             2 & 6.65 & 0.542 & 0.842 \\
             4 & 6.64 & 0.539 & 0.848 \\
             6 & \textbf{6.52} & \textbf{0.531} & \textbf{0.856} \\
             8 & 6.62 & 0.536 & 0.846 \\
             \bottomrule
    \end{tabular}
    }
    \label{tab:ablation_view_nums}
    \vspace{-2mm}
\end{table}

\begin{table}[!tb]
    \centering
    \caption{Percentage of filtered points ($\ell^{1}$ CD ×$10^{3}$ and F-Score@1$\%$).}
    \vspace{-2mm}
    \renewcommand{\arraystretch}{1.0}
    \resizebox{0.8\linewidth}{!}{
        \begin{tabular}{c| c c c}
            \toprule
             Filtered points & CD-Avg$\downarrow$ & DCD-Avg$\downarrow$ & F1$\uparrow$ \\
             \midrule
             12\% & 6.92 & 0.533 & 0.840 \\
             25\% & 6.49 & \textbf{0.518} & \textbf{0.858} \\
             50\% & \textbf{6.47} & 0.530 & 0.857 \\
             \bottomrule
        \end{tabular}
    }
    \vspace{-4mm}
    \label{tab:ablation_filtered_points}
\end{table}

\begin{figure}[H]
    \centering
    \begin{overpic}[width=0.99\linewidth]{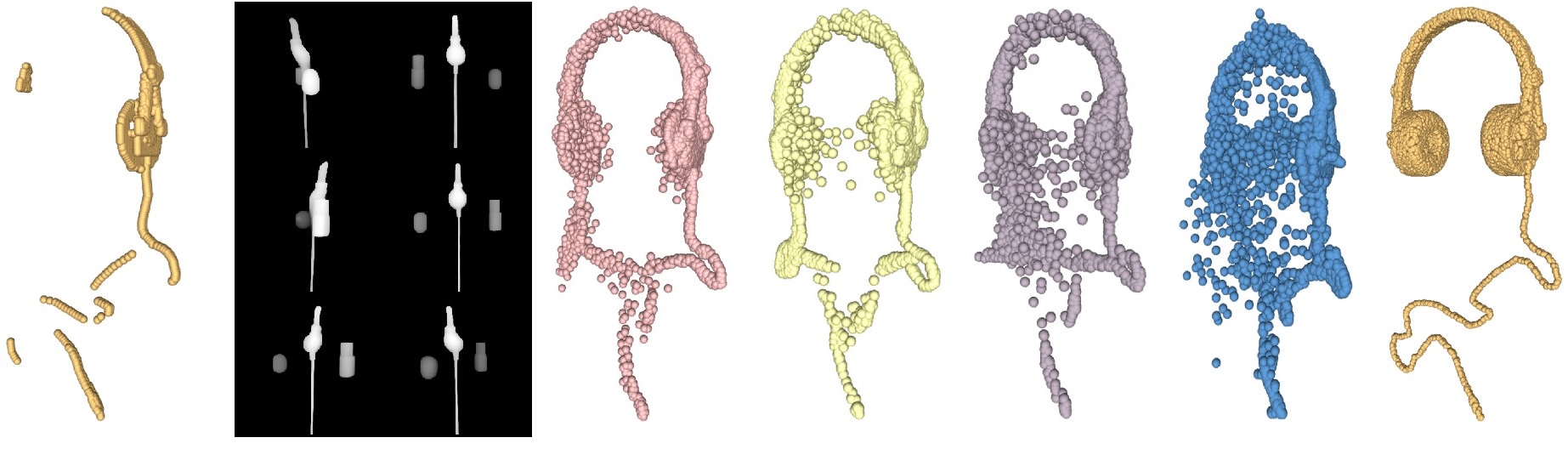}%
        \put(1,-2.5) {\scriptsize (a) Input}
        \put(14,-2.5) {\scriptsize (b) Depth Images}
        \put(37,-2.5) {\scriptsize (c) Ours}
        \put(51,-2.5) {\scriptsize (d) \cite{yu2021pointr}}
        \put(65,-2.5) {\scriptsize (e) \cite{chen2023anchorformer}}
        \put(77,-2.5) {\scriptsize (f) \cite{zhu2023svdformer}}
        \put(90,-2.5) {\scriptsize (g) GT}
    \end{overpic}
    \caption{\textbf{Failure Case.} \revi{Due to the slender structure of the earphone, the multi-view depth images are inconsistent (b), we thus fail to complete the shape. Other methods (d-f) fail as well.}}
    \vspace{-2mm}
    \label{fig:limi}
\end{figure}

\prag{Limitation.} Multi-view depth images can guide point cloud completion, but there are limitations in generating these images. As shown in Fig.~\ref{fig:limi}, the slender structure of the earphone exhibits significant inconsistencies across different generated views, resulting in poor point cloud completion when used as guidance. In cases like this, where the information visible in a single view is extremely limited, the completion results are often unsatisfactory.

\section{Conclusion}
In this paper, we have presented a novel approach for point cloud completion. We introduce diffusion priors in this task to provide imagined full-shape cues for free. From this proof of concept, we believe that diffusion priors can be useful for other point cloud processing tasks, and we hope \mname can inspire future research in this direction.

{
    \small
    \bibliographystyle{ieeenat_fullname}
    \bibliography{main}

\begin{thebibliography}{58}
\providecommand{\natexlab}[1]{#1}
\providecommand{\url}[1]{\texttt{#1}}
\expandafter\ifx\csname urlstyle\endcsname\relax
  \providecommand{\doi}[1]{doi: #1}\else
  \providecommand{\doi}{doi: \begingroup \urlstyle{rm}\Url}\fi

\bibitem[Aiello et~al.(2022)Aiello, Valsesia, and Magli]{aiello2022cross}
Emanuele Aiello, Diego Valsesia, and Enrico Magli.
\newblock Cross-modal learning for image-guided point cloud shape completion.
\newblock \emph{Advances in Neural Information Processing Systems}, 35:\penalty0 37349--37362, 2022.

\bibitem[Blattmann et~al.(2023{\natexlab{a}})Blattmann, Dockhorn, Kulal, Mendelevitch, Kilian, Lorenz, Levi, English, Voleti, Letts, et~al.]{blattmann2023stable}
Andreas Blattmann, Tim Dockhorn, Sumith Kulal, Daniel Mendelevitch, Maciej Kilian, Dominik Lorenz, Yam Levi, Zion English, Vikram Voleti, Adam Letts, et~al.
\newblock Stable video diffusion: Scaling latent video diffusion models to large datasets.
\newblock \emph{arXiv preprint arXiv:2311.15127}, 2023{\natexlab{a}}.

\bibitem[Blattmann et~al.(2023{\natexlab{b}})Blattmann, Rombach, Ling, Dockhorn, Kim, Fidler, and Kreis]{blattmann2023align}
Andreas Blattmann, Robin Rombach, Huan Ling, Tim Dockhorn, Seung~Wook Kim, Sanja Fidler, and Karsten Kreis.
\newblock Align your latents: High-resolution video synthesis with latent diffusion models.
\newblock In \emph{Proceedings of the IEEE/CVF Conference on Computer Vision and Pattern Recognition}, pages 22563--22575, 2023{\natexlab{b}}.

\bibitem[Cadena et~al.(2016)Cadena, Carlone, Carrillo, Latif, Scaramuzza, Neira, Reid, and Leonard]{cadena2016past}
Cesar Cadena, Luca Carlone, Henry Carrillo, Yasir Latif, Davide Scaramuzza, Jos{\'e} Neira, Ian Reid, and John~J Leonard.
\newblock Past, present, and future of simultaneous localization and mapping: Toward the robust-perception age.
\newblock \emph{IEEE Transactions on robotics}, 32\penalty0 (6):\penalty0 1309--1332, 2016.

\bibitem[Chang et~al.(2015)Chang, Funkhouser, Guibas, Hanrahan, Huang, Li, Savarese, Savva, Song, Su, et~al.]{chang2015shapenet}
Angel~X Chang, Thomas Funkhouser, Leonidas Guibas, Pat Hanrahan, Qixing Huang, Zimo Li, Silvio Savarese, Manolis Savva, Shuran Song, Hao Su, et~al.
\newblock Shapenet: An information-rich 3d model repository.
\newblock \emph{arXiv preprint arXiv:1512.03012}, 2015.

\bibitem[Chen et~al.(2023)Chen, Long, Qiu, Yao, Zhou, Luo, and Mei]{chen2023anchorformer}
Zhikai Chen, Fuchen Long, Zhaofan Qiu, Ting Yao, Wengang Zhou, Jiebo Luo, and Tao Mei.
\newblock Anchorformer: Point cloud completion from discriminative nodes.
\newblock In \emph{Proceedings of the IEEE/CVF conference on computer vision and pattern recognition}, pages 13581--13590, 2023.

\bibitem[Dai et~al.(2017)Dai, Ruizhongtai~Qi, and Nie{\ss}ner]{dai2017shape}
Angela Dai, Charles Ruizhongtai~Qi, and Matthias Nie{\ss}ner.
\newblock Shape completion using 3d-encoder-predictor cnns and shape synthesis.
\newblock In \emph{Proceedings of the IEEE conference on computer vision and pattern recognition}, pages 5868--5877, 2017.

\bibitem[Geiger et~al.(2013)Geiger, Lenz, Stiller, and Urtasun]{geiger2013vision}
Andreas Geiger, Philip Lenz, Christoph Stiller, and Raquel Urtasun.
\newblock Vision meets robotics: The kitti dataset.
\newblock \emph{The International Journal of Robotics Research}, 32\penalty0 (11):\penalty0 1231--1237, 2013.

\bibitem[Guo et~al.(2024)Guo, Yang, Rao, Liang, Wang, Qiao, Agrawala, Lin, and Dai]{guo2024animatediffanimatepersonalizedtexttoimage}
Yuwei Guo, Ceyuan Yang, Anyi Rao, Zhengyang Liang, Yaohui Wang, Yu Qiao, Maneesh Agrawala, Dahua Lin, and Bo Dai.
\newblock Animatediff: Animate your personalized text-to-image diffusion models without specific tuning, 2024.

\bibitem[Guo et~al.(2022)Guo, Weng, Fischer, and Fu]{guo20223d}
Yun-Chih Guo, Tzu-Hsuan Weng, Robin Fischer, and Li-Chen Fu.
\newblock 3d semantic segmentation based on spatial-aware convolution and shape completion for augmented reality applications.
\newblock \emph{Computer Vision and Image Understanding}, 224:\penalty0 103550, 2022.

\bibitem[Han et~al.(2017)Han, Li, Huang, Kalogerakis, and Yu]{han2017high}
Xiaoguang Han, Zhen Li, Haibin Huang, Evangelos Kalogerakis, and Yizhou Yu.
\newblock High-resolution shape completion using deep neural networks for global structure and local geometry inference.
\newblock In \emph{Proceedings of the IEEE international conference on computer vision}, pages 85--93, 2017.

\bibitem[He et~al.(2023)He, Yang, Zhang, Shan, and Chen]{he2023latentvideodiffusionmodels}
Yingqing He, Tianyu Yang, Yong Zhang, Ying Shan, and Qifeng Chen.
\newblock Latent video diffusion models for high-fidelity long video generation, 2023.

\bibitem[Hu et~al.(2024)Hu, Zhou, Jampani, and Tulsiani]{hu2024mvd}
Hanzhe Hu, Zhizhuo Zhou, Varun Jampani, and Shubham Tulsiani.
\newblock Mvd-fusion: Single-view 3d via depth-consistent multi-view generation.
\newblock In \emph{Proceedings of the IEEE/CVF Conference on Computer Vision and Pattern Recognition}, pages 9698--9707, 2024.

\bibitem[Hu et~al.(2019)Hu, Han, Shrivastava, and Zwicker]{hu2019render4completion}
Tao Hu, Zhizhong Han, Abhinav Shrivastava, and Matthias Zwicker.
\newblock Render4completion: Synthesizing multi-view depth maps for 3d shape completion.
\newblock In \emph{Proceedings of the IEEE/CVF International Conference on Computer Vision Workshops}, pages 0--0, 2019.

\bibitem[Huang et~al.(2020)Huang, Yu, Xu, Ni, and Le]{huang2020pf}
Zitian Huang, Yikuan Yu, Jiawen Xu, Feng Ni, and Xinyi Le.
\newblock Pf-net: Point fractal network for 3d point cloud completion.
\newblock In \emph{Proceedings of the IEEE/CVF conference on computer vision and pattern recognition}, pages 7662--7670, 2020.

\bibitem[Huang et~al.(2024)Huang, Wen, Dong, Wang, Li, Chen, Cao, Liang, Qiao, Dai, et~al.]{huang2024epidiff}
Zehuan Huang, Hao Wen, Junting Dong, Yaohui Wang, Yangguang Li, Xinyuan Chen, Yan-Pei Cao, Ding Liang, Yu Qiao, Bo Dai, et~al.
\newblock Epidiff: Enhancing multi-view synthesis via localized epipolar-constrained diffusion.
\newblock In \emph{Proceedings of the IEEE/CVF Conference on Computer Vision and Pattern Recognition}, pages 9784--9794, 2024.

\bibitem[Kasten et~al.(2023)Kasten, Rahamim, and Chechik]{kasten2023point}
Yoni Kasten, Ohad Rahamim, and Gal Chechik.
\newblock Point-cloud completion with pretrained text-to-image diffusion models.
\newblock \emph{arXiv preprint arXiv:2306.10533}, 2023.

\bibitem[Kato et~al.(2018)Kato, Tokunaga, Maruyama, Maeda, Hirabayashi, Kitsukawa, Monrroy, Ando, Fujii, and Azumi]{kato2018autoware}
Shinpei Kato, Shota Tokunaga, Yuya Maruyama, Seiya Maeda, Manato Hirabayashi, Yuki Kitsukawa, Abraham Monrroy, Tomohito Ando, Yusuke Fujii, and Takuya Azumi.
\newblock Autoware on board: Enabling autonomous vehicles with embedded systems.
\newblock In \emph{2018 ACM/IEEE 9th International Conference on Cyber-Physical Systems (ICCPS)}, pages 287--296. IEEE, 2018.

\bibitem[Khachatryan et~al.(2023)Khachatryan, Movsisyan, Tadevosyan, Henschel, Wang, Navasardyan, and Shi]{khachatryan2023text2video}
Levon Khachatryan, Andranik Movsisyan, Vahram Tadevosyan, Roberto Henschel, Zhangyang Wang, Shant Navasardyan, and Humphrey Shi.
\newblock Text2video-zero: Text-to-image diffusion models are zero-shot video generators.
\newblock In \emph{Proceedings of the IEEE/CVF International Conference on Computer Vision}, pages 15954--15964, 2023.

\bibitem[Li et~al.(2024{\natexlab{a}})Li, Liu, Long, Zhang, Lin, Li, Qi, Zhang, Luo, Tan, et~al.]{li2024era3d}
Peng Li, Yuan Liu, Xiaoxiao Long, Feihu Zhang, Cheng Lin, Mengfei Li, Xingqun Qi, Shanghang Zhang, Wenhan Luo, Ping Tan, et~al.
\newblock Era3d: High-resolution multiview diffusion using efficient row-wise attention.
\newblock \emph{arXiv preprint arXiv:2405.11616}, 2024{\natexlab{a}}.

\bibitem[Li et~al.(2024{\natexlab{b}})Li, Liu, Chen, Liang, Chen, Tan, and Long]{li2024craftsman}
Weiyu Li, Jiarui Liu, Rui Chen, Yixun Liang, Xuelin Chen, Ping Tan, and Xiaoxiao Long.
\newblock Craftsman: High-fidelity mesh generation with 3d native generation and interactive geometry refiner.
\newblock \emph{arXiv preprint arXiv:2405.14979}, 2024{\natexlab{b}}.

\bibitem[Lin et~al.(2023)Lin, Yue, Hou, Yu, Xu, Yamada, and Zhang]{10378095}
Fangzhou Lin, Yun Yue, Songlin Hou, Xuechu Yu, Yajun Xu, Kazunori~D Yamada, and Ziming Zhang.
\newblock Hyperbolic chamfer distance for point cloud completion.
\newblock In \emph{2023 IEEE/CVF International Conference on Computer Vision (ICCV)}, pages 14549--14560, 2023.

\bibitem[Liu et~al.(2020)Liu, Sheng, Yang, Shao, and Hu]{liu2020morphing}
Minghua Liu, Lu Sheng, Sheng Yang, Jing Shao, and Shi-Min Hu.
\newblock Morphing and sampling network for dense point cloud completion.
\newblock In \emph{Proceedings of the AAAI conference on artificial intelligence}, pages 11596--11603, 2020.

\bibitem[Liu et~al.(2024)Liu, Shi, Chen, Zhang, Xu, Wei, Chen, Zeng, Gu, and Su]{liu2024oneplus}
Minghua Liu, Ruoxi Shi, Linghao Chen, Zhuoyang Zhang, Chao Xu, Xinyue Wei, Hansheng Chen, Chong Zeng, Jiayuan Gu, and Hao Su.
\newblock One-2-3-45++: Fast single image to 3d objects with consistent multi-view generation and 3d diffusion.
\newblock In \emph{Proceedings of the IEEE/CVF Conference on Computer Vision and Pattern Recognition}, pages 10072--10083, 2024.

\bibitem[Liu et~al.(2023)Liu, Lin, Zeng, Long, Liu, Komura, and Wang]{liu2023syncdreamer}
Yuan Liu, Cheng Lin, Zijiao Zeng, Xiaoxiao Long, Lingjie Liu, Taku Komura, and Wenping Wang.
\newblock Syncdreamer: Generating multiview-consistent images from a single-view image.
\newblock \emph{arXiv preprint arXiv:2309.03453}, 2023.

\bibitem[Long et~al.(2024)Long, Guo, Lin, Liu, Dou, Liu, Ma, Zhang, Habermann, Theobalt, et~al.]{long2024wonder3d}
Xiaoxiao Long, Yuan-Chen Guo, Cheng Lin, Yuan Liu, Zhiyang Dou, Lingjie Liu, Yuexin Ma, Song-Hai Zhang, Marc Habermann, Christian Theobalt, et~al.
\newblock Wonder3d: Single image to 3d using cross-domain diffusion.
\newblock In \emph{Proceedings of the IEEE/CVF Conference on Computer Vision and Pattern Recognition}, pages 9970--9980, 2024.

\bibitem[Qi et~al.(2017{\natexlab{a}})Qi, Su, Mo, and Guibas]{qi2017pointnet}
Charles~R Qi, Hao Su, Kaichun Mo, and Leonidas~J Guibas.
\newblock Pointnet: Deep learning on point sets for 3d classification and segmentation.
\newblock In \emph{Proceedings of the IEEE conference on computer vision and pattern recognition}, pages 652--660, 2017{\natexlab{a}}.

\bibitem[Qi et~al.(2017{\natexlab{b}})Qi, Yi, Su, and Guibas]{qi2017pointnet++}
Charles~Ruizhongtai Qi, Li Yi, Hao Su, and Leonidas~J Guibas.
\newblock Pointnet++: Deep hierarchical feature learning on point sets in a metric space.
\newblock \emph{Advances in neural information processing systems}, 30, 2017{\natexlab{b}}.

\bibitem[Qiu et~al.(2024)Qiu, Chen, Gu, Zuo, Xu, Wu, Yuan, Dong, Bo, and Han]{qiu2024richdreamer}
Lingteng Qiu, Guanying Chen, Xiaodong Gu, Qi Zuo, Mutian Xu, Yushuang Wu, Weihao Yuan, Zilong Dong, Liefeng Bo, and Xiaoguang Han.
\newblock Richdreamer: A generalizable normal-depth diffusion model for detail richness in text-to-3d.
\newblock In \emph{Proceedings of the IEEE/CVF Conference on Computer Vision and Pattern Recognition}, pages 9914--9925, 2024.

\bibitem[Reddy et~al.(2018)Reddy, Vo, and Narasimhan]{reddy2018carfusion}
N~Dinesh Reddy, Minh Vo, and Srinivasa~G Narasimhan.
\newblock Carfusion: Combining point tracking and part detection for dynamic 3d reconstruction of vehicles.
\newblock In \emph{Proceedings of the IEEE conference on computer vision and pattern recognition}, pages 1906--1915, 2018.

\bibitem[Rong et~al.(2024)Rong, Zhou, Yuan, Mei, Wang, and Lu]{rong2024cra}
Yi Rong, Haoran Zhou, Lixin Yuan, Cheng Mei, Jiahao Wang, and Tong Lu.
\newblock Cra-pcn: Point cloud completion with intra-and inter-level cross-resolution transformers.
\newblock In \emph{Proceedings of the AAAI Conference on Artificial Intelligence}, pages 4676--4685, 2024.

\bibitem[Shi et~al.(2023)Shi, Wang, Ye, Long, Li, and Yang]{shi2023mvdream}
Yichun Shi, Peng Wang, Jianglong Ye, Mai Long, Kejie Li, and Xiao Yang.
\newblock Mvdream: Multi-view diffusion for 3d generation.
\newblock \emph{arXiv preprint arXiv:2308.16512}, 2023.

\bibitem[Stutz and Geiger(2018)]{stutz2018learning}
David Stutz and Andreas Geiger.
\newblock Learning 3d shape completion from laser scan data with weak supervision.
\newblock In \emph{Proceedings of the IEEE conference on computer vision and pattern recognition}, pages 1955--1964, 2018.

\bibitem[Tang et~al.(2022)Tang, Gong, Yi, Xie, and Ma]{tang2022lake}
Junshu Tang, Zhijun Gong, Ran Yi, Yuan Xie, and Lizhuang Ma.
\newblock Lake-net: Topology-aware point cloud completion by localizing aligned keypoints.
\newblock In \emph{Proceedings of the IEEE/CVF conference on computer vision and pattern recognition}, pages 1726--1735, 2022.

\bibitem[Tang et~al.(2025)Tang, Chen, Chen, Wang, Zeng, and Liu]{tang2025lgm}
Jiaxiang Tang, Zhaoxi Chen, Xiaokang Chen, Tengfei Wang, Gang Zeng, and Ziwei Liu.
\newblock Lgm: Large multi-view gaussian model for high-resolution 3d content creation.
\newblock In \emph{European Conference on Computer Vision}, pages 1--18. Springer, 2025.

\bibitem[Tang et~al.(2023)Tang, Zhang, Chen, Wang, and Furukawa]{NEURIPS2023_a0da690a}
Shitao Tang, Fuyang Zhang, Jiacheng Chen, Peng Wang, and Yasutaka Furukawa.
\newblock Mvdiffusion: Enabling holistic multi-view image generation with correspondence-aware diffusion.
\newblock In \emph{Advances in Neural Information Processing Systems}, pages 51202--51233. Curran Associates, Inc., 2023.

\bibitem[Uy et~al.(2019)Uy, Pham, Hua, Nguyen, and Yeung]{uy-scanobjectnn-iccv19}
Mikaela~Angelina Uy, Quang-Hieu Pham, Binh-Son Hua, Duc~Thanh Nguyen, and Sai-Kit Yeung.
\newblock Revisiting point cloud classification: A new benchmark dataset and classification model on real-world data.
\newblock In \emph{International Conference on Computer Vision (ICCV)}, 2019.

\bibitem[Varley et~al.(2017)Varley, DeChant, Richardson, Ruales, and Allen]{varley2017shape}
Jacob Varley, Chad DeChant, Adam Richardson, Joaqu{\'\i}n Ruales, and Peter Allen.
\newblock Shape completion enabled robotic grasping.
\newblock In \emph{2017 IEEE/RSJ international conference on intelligent robots and systems (IROS)}, pages 2442--2447. IEEE, 2017.

\bibitem[Wang et~al.(2020)Wang, Ang~Jr, and Lee]{wang2020cascaded}
Xiaogang Wang, Marcelo~H Ang~Jr, and Gim~Hee Lee.
\newblock Cascaded refinement network for point cloud completion.
\newblock In \emph{Proceedings of the IEEE/CVF conference on computer vision and pattern recognition}, pages 790--799, 2020.

\bibitem[Wang et~al.(2019)Wang, Sun, Liu, Sarma, Bronstein, and Solomon]{wang2019dynamic}
Yue Wang, Yongbin Sun, Ziwei Liu, Sanjay~E Sarma, Michael~M Bronstein, and Justin~M Solomon.
\newblock Dynamic graph cnn for learning on point clouds.
\newblock \emph{ACM Transactions on Graphics (tog)}, 38\penalty0 (5):\penalty0 1--12, 2019.

\bibitem[Wang et~al.(2024)Wang, Wang, Chen, Xiang, Chen, Yu, Li, Su, and Zhu]{wang2024crm}
Zhengyi Wang, Yikai Wang, Yifei Chen, Chendong Xiang, Shuo Chen, Dajiang Yu, Chongxuan Li, Hang Su, and Jun Zhu.
\newblock Crm: Single image to 3d textured mesh with convolutional reconstruction model.
\newblock \emph{arXiv preprint arXiv:2403.05034}, 2024.

\bibitem[Wen et~al.(2022)Wen, Xiang, Han, Cao, Wan, Zheng, and Liu]{wen2022pmp}
Xin Wen, Peng Xiang, Zhizhong Han, Yan-Pei Cao, Pengfei Wan, Wen Zheng, and Yu-Shen Liu.
\newblock Pmp-net++: Point cloud completion by transformer-enhanced multi-step point moving paths.
\newblock \emph{IEEE Transactions on Pattern Analysis and Machine Intelligence}, 45\penalty0 (1):\penalty0 852--867, 2022.

\bibitem[Wu et~al.(2023)Wu, Zhang, Hou, and Xu]{wu2023leveraging}
Lintai Wu, Qijian Zhang, Junhui Hou, and Yong Xu.
\newblock Leveraging single-view images for unsupervised 3d point cloud completion.
\newblock \emph{IEEE Transactions on Multimedia}, 2023.

\bibitem[Wu et~al.(2021)Wu, Pan, Zhang, Wang, Liu, and Lin]{wu2021balanced}
Tong Wu, Liang Pan, Junzhe Zhang, Tai Wang, Ziwei Liu, and Dahua Lin.
\newblock Balanced chamfer distance as a comprehensive metric for point cloud completion.
\newblock \emph{Advances in Neural Information Processing Systems}, 34:\penalty0 29088--29100, 2021.

\bibitem[Xiang et~al.(2021)Xiang, Wen, Liu, Cao, Wan, Zheng, and Han]{xiang2021snowflakenet}
Peng Xiang, Xin Wen, Yu-Shen Liu, Yan-Pei Cao, Pengfei Wan, Wen Zheng, and Zhizhong Han.
\newblock Snowflakenet: Point cloud completion by snowflake point deconvolution with skip-transformer.
\newblock In \emph{Proceedings of the IEEE/CVF international conference on computer vision}, pages 5499--5509, 2021.

\bibitem[Xie et~al.(2020)Xie, Yao, Zhou, Mao, Zhang, and Sun]{xie2020grnet}
Haozhe Xie, Hongxun Yao, Shangchen Zhou, Jiageng Mao, Shengping Zhang, and Wenxiu Sun.
\newblock Grnet: Gridding residual network for dense point cloud completion.
\newblock In \emph{European conference on computer vision}, pages 365--381. Springer, 2020.

\bibitem[Yan et~al.(2022)Yan, Yan, Wang, Du, Wu, Xie, Pu, and Lu]{yan2022fbnet}
Xuejun Yan, Hongyu Yan, Jingjing Wang, Hang Du, Zhihong Wu, Di Xie, Shiliang Pu, and Li Lu.
\newblock Fbnet: Feedback network for point cloud completion.
\newblock In \emph{European Conference on Computer Vision}, pages 676--693. Springer, 2022.

\bibitem[Yang et~al.(2024)Yang, Kang, Huang, Xu, Feng, and Zhao]{yang2024depth}
Lihe Yang, Bingyi Kang, Zilong Huang, Xiaogang Xu, Jiashi Feng, and Hengshuang Zhao.
\newblock Depth anything: Unleashing the power of large-scale unlabeled data.
\newblock In \emph{Proceedings of the IEEE/CVF Conference on Computer Vision and Pattern Recognition}, pages 10371--10381, 2024.

\bibitem[Yu et~al.()Yu, Huang, Zhang, Li, Tang, and Gao]{yu2024geoformer}
Jinpeng Yu, Binbin Huang, Yuxuan Zhang, Huaxia Li, Xu Tang, and Shenghua Gao.
\newblock Geoformer: Learning point cloud completion with tri-plane integrated transformer.
\newblock In \emph{ACM Multimedia 2024}.

\bibitem[Yu et~al.(2021)Yu, Rao, Wang, Liu, Lu, and Zhou]{yu2021pointr}
Xumin Yu, Yongming Rao, Ziyi Wang, Zuyan Liu, Jiwen Lu, and Jie Zhou.
\newblock Pointr: Diverse point cloud completion with geometry-aware transformers.
\newblock In \emph{Proceedings of the IEEE/CVF international conference on computer vision}, pages 12498--12507, 2021.

\bibitem[Yuan et~al.(2018)Yuan, Khot, Held, Mertz, and Hebert]{yuan2018pcn}
Wentao Yuan, Tejas Khot, David Held, Christoph Mertz, and Martial Hebert.
\newblock Pcn: Point completion network.
\newblock In \emph{2018 international conference on 3D vision (3DV)}, pages 728--737. IEEE, 2018.

\bibitem[Zhang et~al.(2023{\natexlab{a}})Zhang, Rao, and Agrawala]{zhang2023adding}
Lvmin Zhang, Anyi Rao, and Maneesh Agrawala.
\newblock Adding conditional control to text-to-image diffusion models.
\newblock In \emph{Proceedings of the IEEE/CVF International Conference on Computer Vision}, pages 3836--3847, 2023{\natexlab{a}}.

\bibitem[Zhang et~al.(2021)Zhang, Feng, Li, Zou, Wan, Zhao, Guo, and Gao]{zhang2021view}
Xuancheng Zhang, Yutong Feng, Siqi Li, Changqing Zou, Hai Wan, Xibin Zhao, Yandong Guo, and Yue Gao.
\newblock View-guided point cloud completion.
\newblock In \emph{Proceedings of the IEEE/CVF conference on computer vision and pattern recognition}, pages 15890--15899, 2021.

\bibitem[Zhang et~al.(2023{\natexlab{b}})Zhang, Wei, Jiang, Zhang, Zuo, and Tian]{zhang2023controlvideo}
Yabo Zhang, Yuxiang Wei, Dongsheng Jiang, Xiaopeng Zhang, Wangmeng Zuo, and Qi Tian.
\newblock Controlvideo: Training-free controllable text-to-video generation.
\newblock \emph{arXiv preprint arXiv:2305.13077}, 2023{\natexlab{b}}.

\bibitem[Zhao et~al.(2021)Zhao, Jiang, Jia, Torr, and Koltun]{zhao2021point}
Hengshuang Zhao, Li Jiang, Jiaya Jia, Philip~HS Torr, and Vladlen Koltun.
\newblock Point transformer.
\newblock In \emph{Proceedings of the IEEE/CVF international conference on computer vision}, pages 16259--16268, 2021.

\bibitem[Zhou et~al.(2022{\natexlab{a}})Zhou, Wang, Yan, Lv, Zhu, and Feng]{zhou2022magicvideo}
Daquan Zhou, Weimin Wang, Hanshu Yan, Weiwei Lv, Yizhe Zhu, and Jiashi Feng.
\newblock Magicvideo: Efficient video generation with latent diffusion models.
\newblock \emph{arXiv preprint arXiv:2211.11018}, 2022{\natexlab{a}}.

\bibitem[Zhou et~al.(2022{\natexlab{b}})Zhou, Cao, Chu, Zhu, Lu, Tai, and Wang]{zhou2022seedformer}
Haoran Zhou, Yun Cao, Wenqing Chu, Junwei Zhu, Tong Lu, Ying Tai, and Chengjie Wang.
\newblock Seedformer: Patch seeds based point cloud completion with upsample transformer.
\newblock In \emph{European conference on computer vision}, pages 416--432. Springer, 2022{\natexlab{b}}.

\bibitem[Zhu et~al.(2023)Zhu, Chen, He, Wang, Qin, and Wei]{zhu2023svdformer}
Zhe Zhu, Honghua Chen, Xing He, Weiming Wang, Jing Qin, and Mingqiang Wei.
\newblock Svdformer: Complementing point cloud via self-view augmentation and self-structure dual-generator.
\newblock In \emph{Proceedings of the IEEE/CVF International Conference on Computer Vision}, pages 14508--14518, 2023.

\end{thebibliography}
}

\clearpage
\appendix
\setcounter{page}{1}

\setcounter{table}{0}
\renewcommand{\thetable}{A\arabic{table}}
\setcounter{figure}{0}
\renewcommand{\thefigure}{A\arabic{figure}}

\maketitlesupplementary

In this supplementary, we present \revi{experiments on additional datasets (\ie, ScanObjectNN~\cite{uy-scanobjectnn-iccv19}, KITTI~\cite{geiger2013vision}, ShapeNet-55~\cite{yu2021pointr}) to demonstrate the generalization ability of our method on real-world scans, as well as randomly cropped general partial points clouds. 
Besides, we show complementary visual results on PCN~\cite{yuan2018pcn}  and ShapeNet-55~\cite{yu2021pointr} datasets and present additional visual examples when comparing our approach against SDS-Complete~\cite{kasten2023point} and other single-view 3D generation methods.}

\begin{figure}[!htb]
    \begin{overpic}[width=0.99\linewidth]{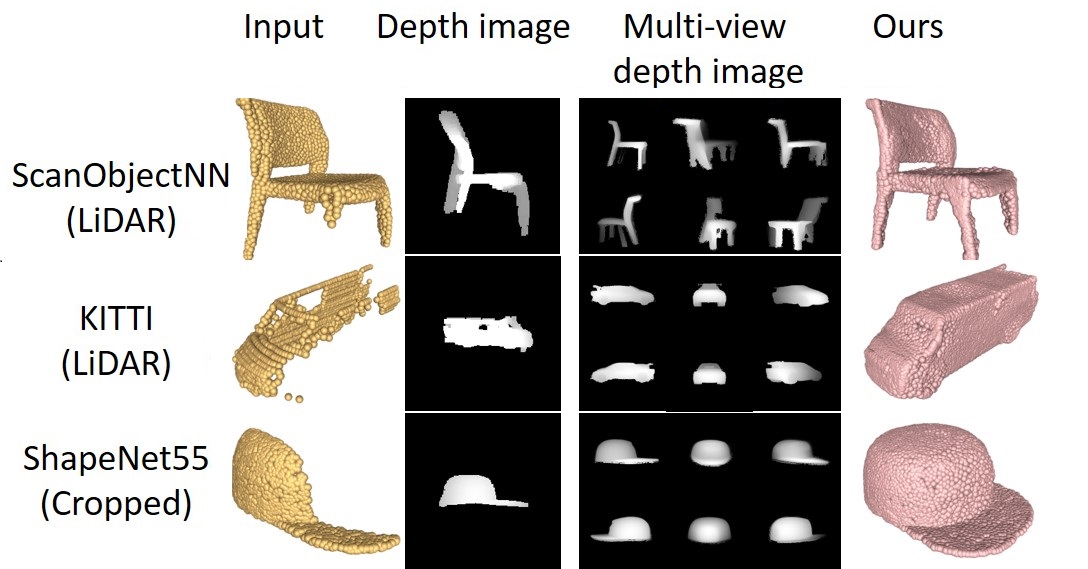}
    \end{overpic}   
    \caption{{Visual results on real scanned and randomly cropped datasets.}
    The first and second rows are real scanned data from ScanObjectNN and KITTI, while the third row is randomly cropped models from ShapeNet55. The converted single-view depth image and generated multi-view depth images are shown in columns two and three.}
    \vspace{-4mm}
    \label{fig:scans_rebuttle}
\end{figure}

\revi{
\paragraph{Real-world scans and general partial point clouds.}
As stated in \cref{sec:intro} in the paper, completing a single-view partial point cloud is actively studied and much more challenging than completing a general one, since it usually misses more than half of the points (see the chair and lamp in \cref{fig:teaser} in the main paper).
Due to the unique setting, we thus design our \mname, dedicated to completing a single-view partial point cloud faithfully. 
However, thanks to the robustness of the large diffusion models and our fuser and consolidator, \mname can work on general partial clouds well.
Following the convention, we did not emphasize and report this advantage in the paper, but preliminary visual results are presented in Fig. \ref{fig:scans_rebuttle}, where \mname successfully completes both LiDAR point clouds and randomly cropped ShapeNet55 point clouds.
Indeed, a perfect single-view depth image is not feasible in this case, we thus select the most informative view and obtain the `incomplete' depth image serving as the input of our method. To mitigate the incompleteness and potential quality degradation in the initial depth map, we employ a larger guidance scale (\eg, 9.0) along with more complex text prompts in ControlNet. This strategy effectively enhances the quality of the generated RGB images.
Extending our method to handle arbitrary multi-view depth images is a promising direction, we leave it for future work.
}

\begin{figure}[!t]
    \centering
    \includegraphics[width=0.95\linewidth]{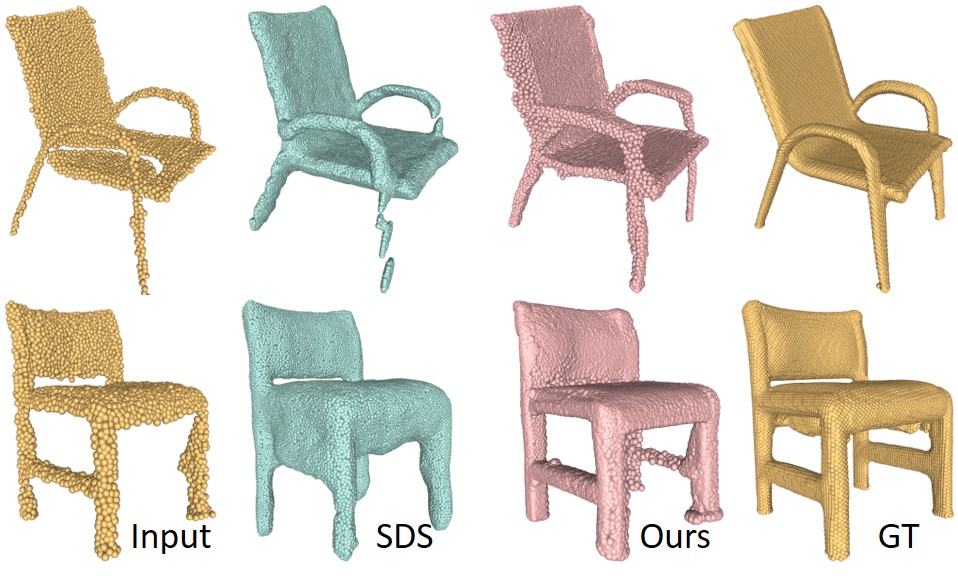}
    \vspace{-3mm}
    \caption{The visual comparison with SDS-Complete.}
    \vspace{-2mm}
    \label{fig:sds}
\end{figure}

\revi{
\paragraph{Comparison with SDS-Complete~\cite{kasten2023point}.}
We have conducted a visual comparison with SDS-Complete, which exploits diffusion priors with SDS optimization. The results in Fig. \ref{fig:sds} demonstrate that while SDS-Complete is capable of recovering the overall shape, it fails to capture fine details, such as the leg and the armrest of the chair.
}

\begin{figure*}[!ht]
    \centering
    \begin{overpic}[width=\textwidth]{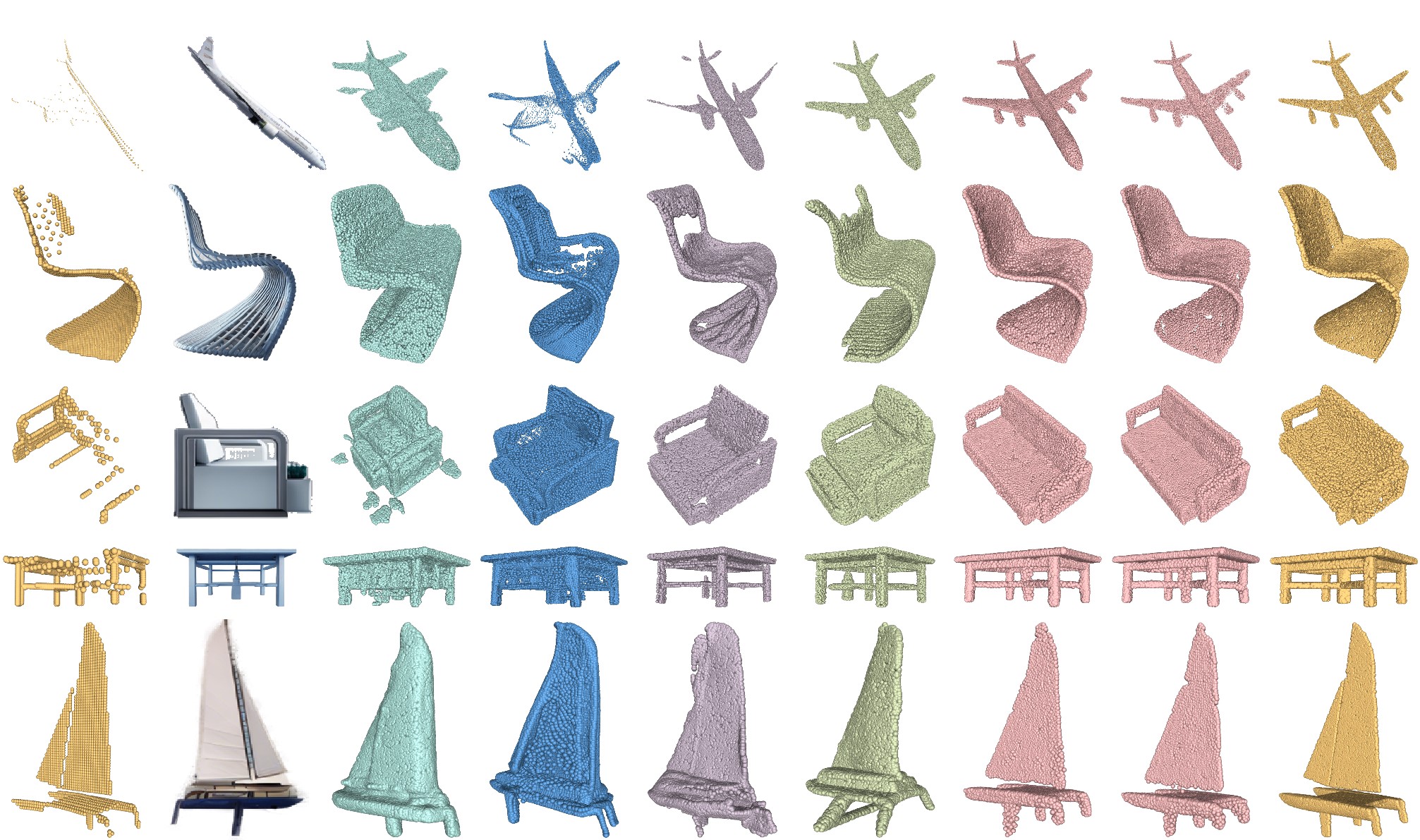}
        \put(5,-3.5) {\scriptsize Input}
        \put(15, -3.5) {\scriptsize \shortstack{RGB\\image}}
        \put(25,-3.5) {\scriptsize CRM~\cite{wang2024crm}}
        \put(35,-3.5) {\scriptsize LGM~\cite{tang2025lgm}}
        \put(45, -3.5) {\scriptsize One-2-3-45++~\cite{liu2024oneplus}}
        \put(58, -3.5) {\scriptsize Craftsman~\cite{li2024craftsman}}
        \put(70,-3.5) {\scriptsize \shortstack{Ours\\(SVD-MV)}}
        \put(80,-3.5) {\scriptsize \shortstack{Ours\\(Wonder3D)}}
        \put(94,-3.5) {\scriptsize GT}
    \end{overpic}   
    \vspace{0.5mm}
    \caption{Additional visual comparisons with 3D model generation baesd on single-view RGB image.}
    \vspace{-3mm}
    \label{fig:com_recon_supp}
\end{figure*}

\paragraph{Additional Comparison with 3D generation methods.}
In \cref{fig:com_recon_supp}, we present additional comparisons with other 3D model generation methods~\cite{wang2024crm, tang2025lgm}. Moreover, we provide further examples from the PCN dataset for comprehensive evaluation. As illustrated in the figure, the lack of extra views, the randomness in the generation process, and the inherent inconsistencies in multi-view images often prevent single-view 3D model generation methods from effectively capturing geometric details. Consequently, these models may produce shapes that, while plausible, exhibit significant deviations from the ground truth, such as variations in the width of the sofa. Therefore, multi-view images are more appropriate as auxiliary cues for point cloud completion, although directly utilizing them for point cloud completion typically yields suboptimal results.

\begin{figure*}[!htb]
    \centering
    \begin{overpic}[width=\textwidth]{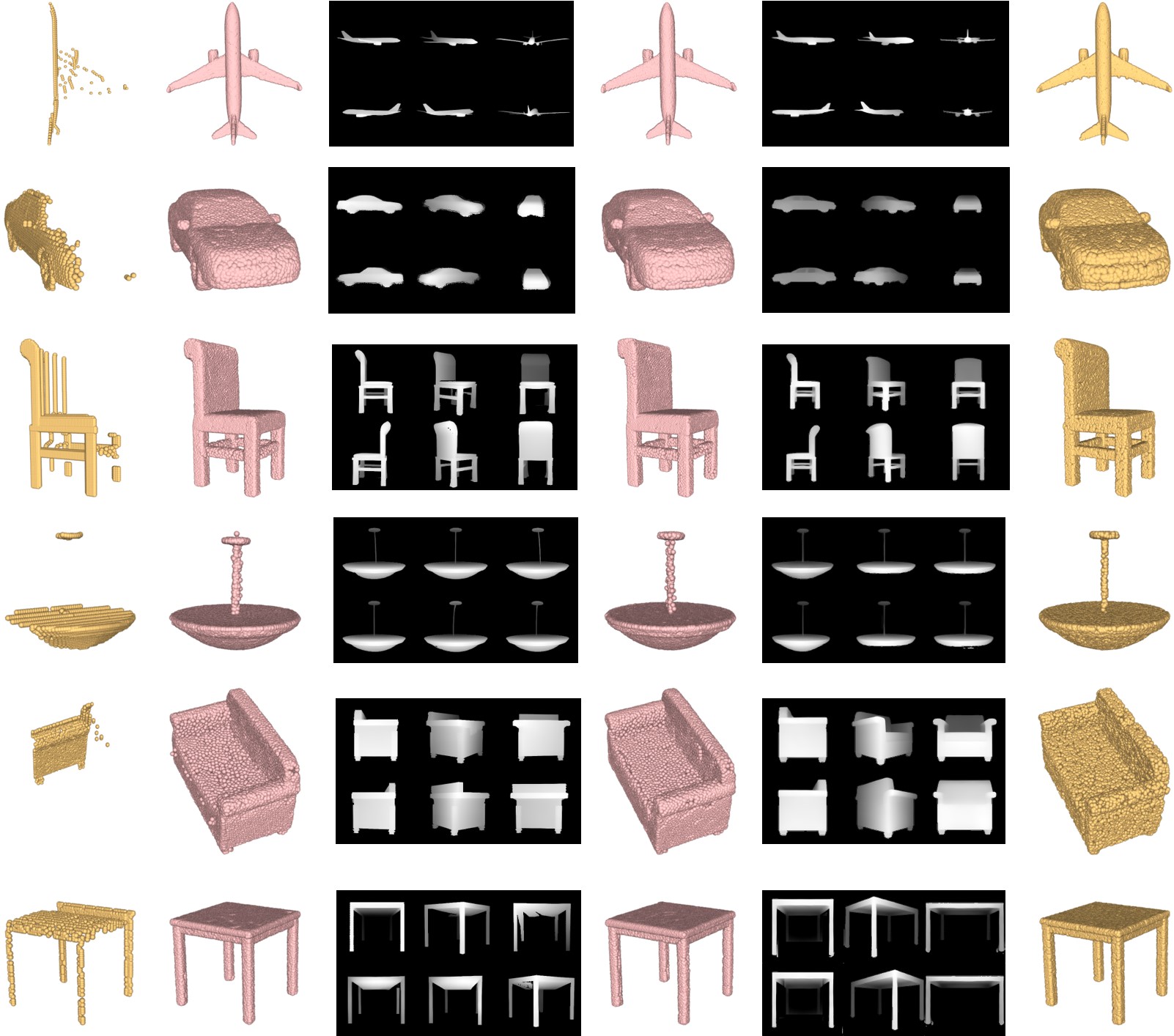}
        \put(5,-4.5) {\scriptsize{Input}}
        \put(18,-4.5) {\scriptsize{\shortstack{Ours\\(SVD-MV)}}}
        \put(34,-4.5) {\scriptsize{\shortstack{Depth images\\(SVD-MV)}}}
        \put(53,-4.5) {\scriptsize{\shortstack{Ours\\(Wonder3D)}}}
        \put(70,-4.5) {\scriptsize{\shortstack{Depth images\\(Wonder3D)}}}
        \put(93,-4.5) {GT}
    \end{overpic}   
    \vspace{2mm}
    \caption{\textbf{Visual results as well as multi-view depth images on the PCN dataset.} The generated multi-view depth images corresponding to the last three rows exhibit lower quality and less consistency than the first three rows.}
    \label{fig:pcn_supp}
\end{figure*}

\begin{figure*}[!htb]
    \centering
    \begin{overpic}[width=\textwidth]{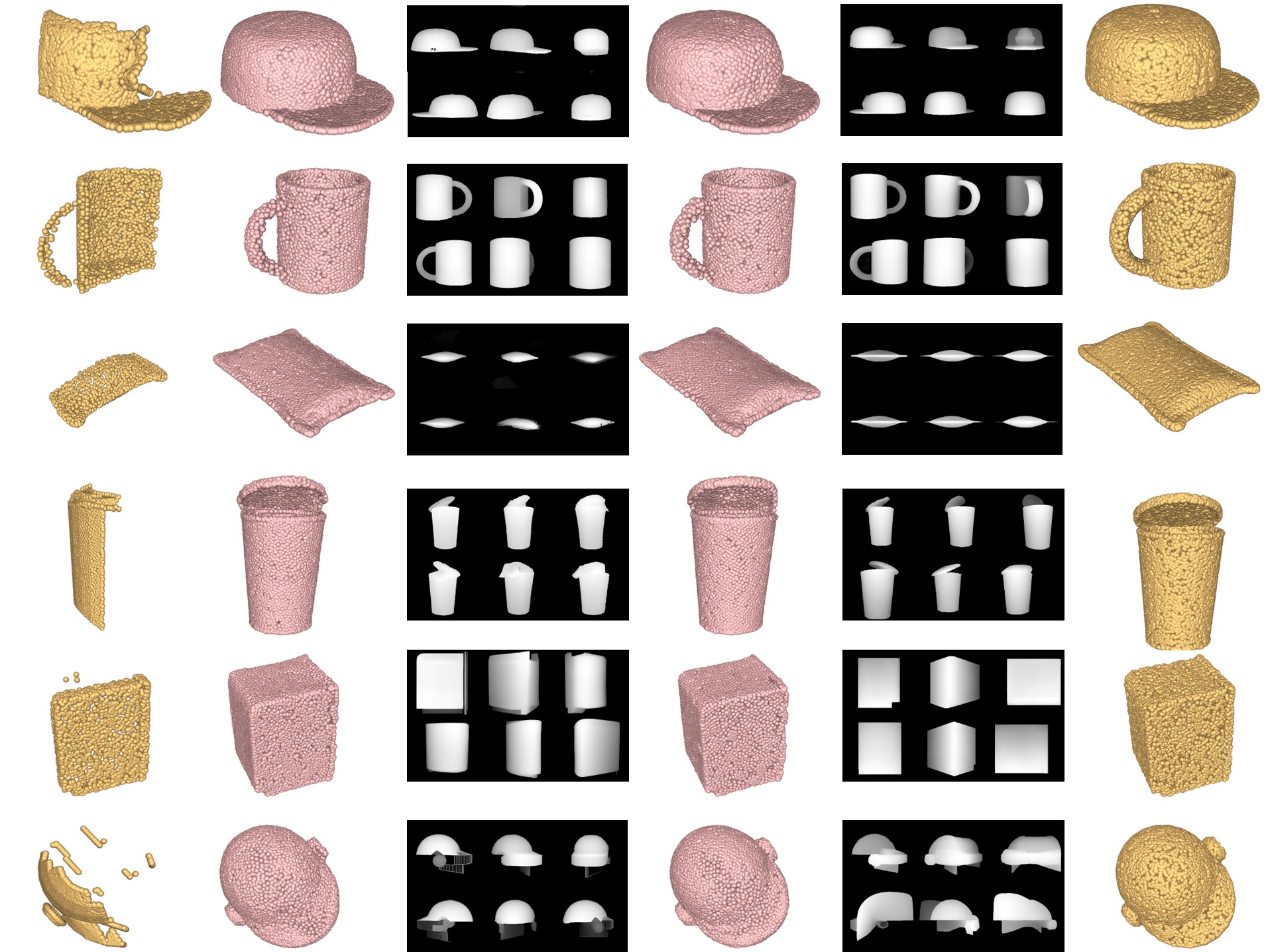}
        \put(5,-4.5) {\scriptsize{Input}}
        \put(18,-4.5) {\scriptsize{\shortstack{Ours\\(SVD-MV)}}}
        \put(35,-4.5) {\scriptsize{\shortstack{Depth images\\(SVD-MV)}}}
        \put(53,-4.5) {\scriptsize{\shortstack{Ours\\(Wonder3D)}}}
        \put(70,-4.5) {\scriptsize{\shortstack{Depth images\\(Wonder3D)}}}
        \put(93,-4.5) {GT}
    \end{overpic}   
    \vspace{2mm}
    \caption{\textbf{Visual results as well as multi-view depth images on the ShapeNet-55 dataset.} The generated multi-view depth images corresponding to the last three rows exhibit lower quality and less consistency than the first three rows.}
    \label{fig:shapenet55_supp}
\end{figure*}

\paragraph{Additional results on PCN and ShapeNet-55 dataset.}
\cref{fig:pcn_supp,fig:shapenet55_supp}, each presents an extra six results from the PCN and the ShapeNet-55 datasets, respectively. 
For both datasets, the first three cases in the first three rows possess high-quality and consistent multi-view depth images, whereas depth images with lower quality and less consistency are generated for the last three examples. 
Specifically, the high randomness of the generation process, combined with limitations of the initial view, can result in multi-view depth images exhibiting inconsistencies across views (\eg, the lid of the \textit{Dustbin}), missing geometric details(\eg, the \textit{Helmet}), and the presence of noisy regions (\eg, the leg of the \textit{Table}). Furthermore, shapes imagined by the generative model may exhibit scale discrepancies relative to their real-world counterparts (\eg, the \textit{Sofa} and the \textit{Table}). 
In addition, the depth estimation process may also occasionally produce suboptimal outcomes (\eg, the \textit{Dishwasher}). 
However, the proposed confidence-based shape consolidator effectively addresses these issues by eliminating unreliable points caused by inconsistencies in diffusion priors. As a result, our models produce accurate and reasonable outputs, as demonstrated in the second and fourth columns of the last three examples.

\end{document}